\documentclass[a4paper]{iopart}
\usepackage[utf8]{inputenc}
\usepackage[T1]{fontenc}
\usepackage{gensymb,eurosym,url,soul} 
\usepackage{caption}
\usepackage[font=footnotesize,textfont=footnotesize]{subcaption}
\usepackage{longtable,booktabs,array}
\usepackage[backend=biber, sorting=none, style=numeric-comp, maxnames=5]{biblatex}
\usepackage[switch]{lineno}
\usepackage{multirow}
\usepackage{listings}
\usepackage{acronym}
\usepackage{graphicx}
\usepackage{xcolor}
\usepackage{xr} 
\usepackage{tikz}
\usepackage[europeanresistors]{circuitikz}
\usepackage{ifthen}
\usepackage[perpage, symbol]{footmisc}
\usepackage{hyperref}

\hypersetup{colorlinks=true, breaklinks=true, 
  urlcolor=blue, linkcolor=blue,anchorcolor=blue,citecolor=blue,
  hypertexnames=true, final=true, 
  pdfpagemode = UseNone, 
  pdfauthor = {},
  pdftitle = {},   
  pdfsubject = {},
  pdfkeywords = {}
}

\graphicspath{{img/} {./} {figures/}}
\DeclareGraphicsExtensions{.pdf,.png,.jpg,.mps}

\DeclareSourcemap{
  \maps[datatype=bibtex]{
    \map[overwrite=true]{
      \step[fieldset=url, null]
      \step[fieldset=eprint, null]
    }
  }
}

\usetikzlibrary{arrows.meta, decorations.pathreplacing, fit, positioning, shapes.arrows, shapes.multipart, shapes.geometric}
\tikzstyle{startstop} = [rectangle, rounded corners, text centered, draw=black]
\tikzstyle{io} = [trapezium, trapezium left angle=70, trapezium right angle=110, text centered, draw=black]
\tikzstyle{process} = [rectangle, text centered, draw=black]
\tikzstyle{decision} = [diamond, aspect=2, text centered, draw=black]
\tikzstyle{arrow} = [thick,->,>=stealth]
\ctikzset{tripoles/pmos style=emptycircle}

\newcommand{\chipname}[0]{DYNAP-SE2}
\newcommand{\softwarename}[0]{Samna}

\newcommand{\companyname}[0]{Synsense}

\begin{document}

\title[DYNAP-SE2]{DYNAP-SE2: a scalable multi-core dynamic neuromorphic asynchronous spiking neural network processor}
\author{Ole Richter $^{1,3,4,*}$, Chenxi Wu$^{2,*}$, Adrian M. Whatley$^{2}$, German K\"ostinger$^{2}$, Carsten Nielsen$^{1,2}$, Ning Qiao$^{1,2}$, Giacomo Indiveri$^{2}$}

\address{$^{1}$SynSense AG, Zurich, Switzerland}
\address{$^{2}$Institute of Neuroinformatics, University of Z\"urich and ETH Z\"urich, Switzerland}
\address{$^{3}$Zernike Institute for Advanced Materials, University of Groningen, Netherlands}
\address{$^{4}$Groningen Cognitive Systems and Materials Center (CogniGron), University of Groningen, Netherlands}
\address{$^{*}$both authors contributed equally}
\ead{giacomo@ini.uzh.ch, chenxi@ini.uzh.ch, o.j.richter@rug.nl}


\begin{abstract}
With the remarkable progress that technology has made, the need for processing data near the sensors at the edge has increased dramatically.
The electronic systems used in these applications must process data continuously, in real-time, and extract relevant information using the smallest possible energy budgets.
A promising approach for implementing always-on processing of sensory signals that supports on-demand, sparse, and edge-computing is to take inspiration from biological nervous system.
Following this approach, we present a brain-inspired platform for prototyping real-time event-based Spiking Neural Networks (SNNs).
The system proposed supports the direct emulation of dynamic and realistic neural processing phenomena such as short-term plasticity, NMDA gating, AMPA diffusion, homeostasis, spike frequency adaptation, conductance-based dendritic compartments and spike transmission delays.
The analog circuits that implement such primitives are paired with a low latency asynchronous digital circuits for routing and mapping events.
This asynchronous infrastructure enables the definition of different network architectures, and provides direct event-based interfaces to convert and encode data from event-based and continuous-signal sensors. 
Here we describe the overall system architecture, we characterize the mixed signal analog-digital circuits that emulate neural dynamics, demonstrate their features with experimental measurements, and present a low- and high-level software ecosystem that can be used for configuring the system. 
The flexibility to emulate different biologically plausible neural networks, and the chip's ability to monitor both population and single neuron signals in real-time, allow to develop and validate complex models of neural processing for both basic research and edge-computing applications.  
\end{abstract}

\maketitle

\section{Introduction}
\begin{figure}[b]
  \centering
  \includegraphics[width=0.6\textwidth]{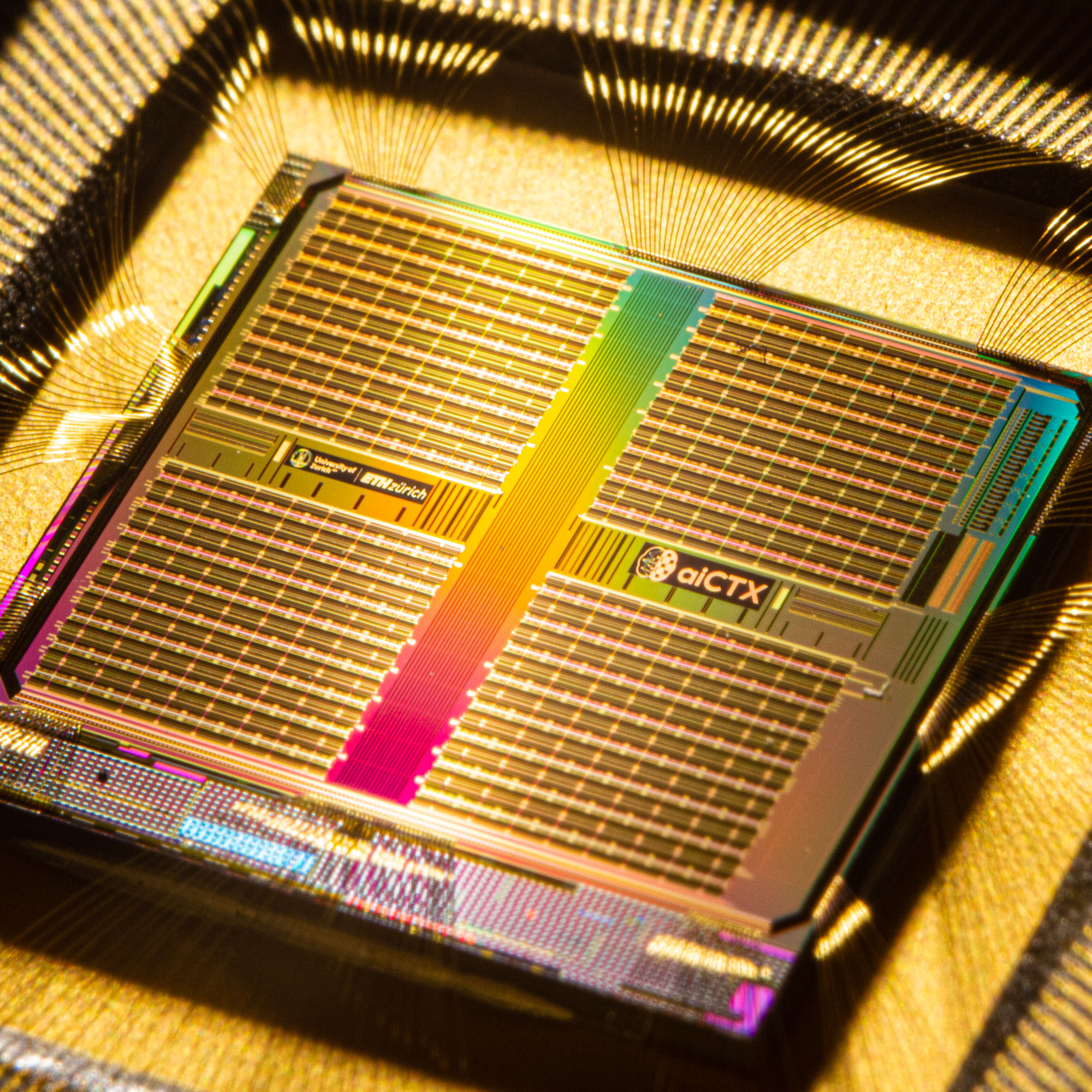}
  \caption{Photo of the \chipname{} chip, which has an area of $98\,mm^2$ manufactured in 180nm CMOS technology as a cost effective prototyping platform.}
  \label{fig:photo-of-the-\chipname{}-chip}
\end{figure}

As technology has progressed, the need for processing more sensory data at the edge has increased dramatically.
In particular, an increasing amount of applications are expected to process data near the sensors, without resorting to remote computing servers.
For these types of applications it is of prime importance to minimize power consumption and latency, while maintaining robustness and adaptability to changing conditions.
The processors used in these applications therefore need to process the data being measured by the sensors continuously, in real-time, and to extract relevant information using the smallest possible energy budgets.
A promising approach for implementing always-on processing of sensory signals that supports on-demand, sparse, and edge-intelligence computation, is that of using event-based Spiking Neural Networks (SNNs)~\cite{Liu_etal14,Gallego_etal20,Mostafa_etal15a,Corradi_etal14,Maass97,Kasabov_etal13,Huynh_etal22}.
The event-based representation has been shown to be particularly well suited to transmitting analog signals across noisy channels, while maximizing robustness to noise and minimizing bandwidth requirements and power consumption~\cite{Liu_etal14,Lazzaro_etal93,Boahen98}.
Furthermore, by encoding only the changes in the signals, this representation is optimally suited for sensory signals that change sparsely in time, producing data only when necessary~\cite{Gallego_etal20,Corradi_etal14}.
The computational paradigm that best exploits the event-based representation is that of SNNs.

As one of the largest sources of energy consumption in electronic processing systems is \emph{data-movement}~\cite{Mittal18,Mutlu_etal19}, the best way to minimize power consumption in event-based SNNs is to implement them as massively-parallel in-memory computing architectures that process the data on the fly, as it is being sensed, without having to store it and retrieve it.
It is therefore important to match the rate of the data arriving in input to the processing rate and the time constants of the synapses and neurons in the SNN.\@
Neuron and synapse circuits can be configured to process natural signals such as human voice, gestures, or bio-signals, by setting their time constants to tens or hundreds of milliseconds (and significantly reducing their processing speed). This can improve the information retention and processing ability of feed--forward SNNs. 
However, processing of signals that contain very long and multiple timescales using this approach requires resorting to recurrent SNNs (RNNs)~\cite{Indiveri_Sandamirskaya19,Sussillo14,Amit_etal85}.
These types of networks provide a valuable algorithmic foundation for adaptive and efficient processing of continuous sensory signals, as they can be configured to exhibit a wide range of dynamics that are fundamental in lowering the amount of storage resources required to process, recognize, and generate long temporal sequences and patterns.

Conventional neural network accelerators and digital implementations of SNNs~\cite{Furber_Bogdan20,Davies_etal18} can be in principle used to design and train both feed-forward and recurrent neural networks.
However their memory storage and data movement requirements increase their power budget significantly and negates their advantages compared to using standard computing architectures~\cite{Knight_Nowotny18}.
The original neuromorphic engineering approach proposed in~\cite{Mead20,Chicca_etal14} aims to solve the above challenges by using analog circuits that operate in weak-inversion (subthreshold) and in physical time to implement neural dynamics for solving sensory processing tasks, in a data-driven manner.
In this approach each neuron and synapse computational element is implemented using a dedicated physical circuit, without resorting to time-multiplexing of shared computing resources.
Computation is therefore massively parallel and distributed, and takes place only if the synapse/neuron is driven by input events. 
For interactive real world data processing, the event-based mixed signal approach is an optimal match: it allows carrying out physical-time sensory processing with low-power circuits, and the implementation of artificial intelligence computing primitives for solving extreme edge computing applications~\cite{Indiveri_Sandamirskaya19,Chicca_etal14}.

In this paper we present a mixed-signal neuromorphic processor that follows this approach.
It directly emulates the dynamics of biological neurons and synapses using analog integrated circuits for computation, and asynchronous digital circuits for transmitting the events (spikes) produced by the neurons to destination synapses or to the output pads. The processor features a clock-free asynchronous digital hierarchical routing scheme which runs in native real-time and ensures low latency~
\cite{Moradi_etal18}.
The processor we present is denoted as the DYnamic Neuromorphic Asynchronous Processor-ScalablE 2 
\chipname{}
This chip significantly extends the features of the previous generation DYNAP-SE~\cite{Moradi_etal18} at the synapse and neuron circuit level, at the network-level, and at the asynchronous routing fabric level.
We show here how the \chipname{} offers rich neuronal dynamics across different timescales to support a wide spectrum of biologically plausible recurrent networks.
We present the overall architecture and describe in detail the individual circuits, providing experimental results measured from the chip to validate the theory.
To enable near-sensor processing the \chipname{} also integrates an on-chip analog front-end (AFE) with low-noise amplifiers, band-pass filters and asynchronous delta modulators for converting input waveforms into streams of address-events
~\cite{Sharifshazileh_etal21}.
Similarly, \chipname{} includes a direct 2D sensor event pre-processor
\cite{Richter_etal23}
that can cut, scale and arbitrarily map 2D event stimuli from a Dynamic Vision Sensor (DVS)~\cite{Lichtsteiner_etal08}.


The structure of this paper is the following. Section \ref{sec:chip-overview} presents an overview of the general architecture and available resources of the chip. Section \ref{sec:common-neuromorphic-building-blocks} reviews the common building blocks that are crucial to understanding and using the chip. Section \ref{sec:analog-neuronal-circuits} enumerates the core analog neural circuit with application examples and real measurement. Section \ref{sec:digital-spike-routing-scheme} elaborates the routing scheme and methods for building large scale neural networks. Section \ref{sec:interfaces} briefly describes the interfaces the chip presents to the outside world and Sec.~\ref{sec:software} describes the software system that supports the usability of the chip. As the analog front-end is independent of the neuron cores and event processing, for more information regarding its circuit design and application see 
\cite{Sharifshazileh_etal21}.



\section{Chip overview} \label{sec:chip-overview}
\subsection{System architecture}

Computation is centered on the 1024, analog, integrate-and-fire neurons arranged in $2\times2$ cores of grids of $16\times16$ neurons each. Each neuron has 64 synapses and four dendritic branches. The only way to send information to the neurons and for the neurons to send information out is through digital spikes. The routing scheme will be elaborated in Section \ref{sec:digital-spike-routing-scheme}.
As opposed to many computational models, the neurons do not receive analog current injection directly, and the membrane potential is also not accessible.
These design choices are taken for scalability reasons, because there is no easy way to access thousands of analog values at the same time, while digitized spikes can easily be routed using time-multiplexing \cite{Lazzaro_Wawrzynek95}.
Thus, in order to provide analog input to the network, a neuromorphic sensor \cite{Liu_Delbruck10} (such as a DVS \cite{Delbruck_Mead94} or AFE 
\cite{Sharifshazileh_etal21}
) that encodes a signal into spikes is needed, and the computation and learning algorithms should be completely spike-based.

\begin{figure}[!htbp]
    \centering
    \scalebox{0.8}{
    \begin{tikzpicture}
        
        \foreach \y in {0, 1.5, 3, 4.5} {
            \draw[-latex] (-6.25,\y) -- (-1.5, \y);
            \draw (-1.5,\y-0.5) rectangle (0.25,\y+0.5);
        }
        \foreach \i in {1, 2, ..., 64}{
            \pgfmathsetmacro\y{{int((rand+1)*2)*1.5}}
            \pgfmathsetmacro\x{{(rand)*2-3.5}}
            \draw[-latex] (\x - 0.5, \y - 0.75) -- (\x - 0.5, \y);
        }
        \foreach \y in {3, 4.5} \node at(-0.625, \y) {(D)DPI};
        \foreach \y in {0, 1.5} \node at(-0.625, \y) {DPI};
        \draw (0.25,4) rectangle (3,5);
        \node at(-0.625, 5.25) {AMPA};
        \node at(1.625, 4.75) {(resistive grid)};
        \node at(1.625, 4.25) {(conductance)};
        \draw (0.25,2.5) rectangle (3,3.5);
        \node at(-0.625, 3.75) {NMDA};
        \node at(1.625, 3.25) {($V_{mem}$ gating)};
        \node at(1.625, 2.75) {(conductance)};
        \draw (0.25,1) rectangle (3,2);
        \node at(-0.625, 2.25) {GABA B};
        \node at(1.625, 1.5) {(conductance)};
        \node at(-0.625, 0.75) {GABA A};
        \node[draw, circle] at (4, 3) {+};
        \draw[-latex] (3, 1.5) -| (4, 2.6) node[below right] {$-$};
        \draw[-latex] (3, 3) -- (3.6, 3) node[above left] {$+$};
        \draw[-latex] (3, 4.5) -| (4, 3.4) node[above right] {$+$};
        \draw[-latex] (4.4, 3) -- (5, 3);
        \draw (5, 3) -- (6.732, 4) -- (6.732, 2) -- (5, 3);
        \draw[-latex] (0.25, 0) -- (4,0) -- (5.866, 2.5) node[left] {$-$};
        \draw (6.732, 3) -- (7.366,3);
        \draw[-latex] (7.366, 2.5) |- (9,1.5);
        \draw[-latex] (7.366, 3) |- (9,2.5);
        \draw[-latex] (7.366, 3) |- (9,3.5);
        \draw[-latex] (7.366, 3.5) |- (9,4.5);
        \node at (-3.75, 6.5) {Synapse $\times$ 64};
        \node at (-3.75, 6) {4-bit weight};
        \node at (-3.75, 5.5) {(delay)};
        \node at (-3.75, 5) {(STP)};
        \node at (0.75, 6.5) {Dendritic branches};
        \node at (5.866, 6.5) {Soma};
        \node at (5.866, 6) {exp/thr};
        \node at (5.866, 5.5) {`calcium'};
        \node at (5.866, 5) {(adaptation)};
        \node at (8, 6.5) {Axon $\times$ 4};
        \node at (8, 6) {target chip};
        \node at (8, 5.5) {core mask};
        \node at (8, 5) {11-bit tag};
    \end{tikzpicture}}
    \caption{Neuronal compartments. 64 synapses with 4-bit weights and conditional delay and short-term plasticity (STP) convert pre-synaptic spikes to pulses. The pulses are low-pass-filtered by one of the four dendrites to generate post-synaptic currents (PSC). The dendrites have conditional alpha-function excitatory PSCs, a diffusive grid, membrane voltage gating and ion-channel conductances. The PSCs are injected into the soma, which can switch between a thresholded \cite{Rubino_etal19} and exponential integrate-and-fire model \cite{Brette_Gerstner05}, with conditional adaptation and `calcium'-based homeostasis. When the neuron fires, the AER spike is sent to up to four chips.} 
    \label{fig:neuron-model}
\end{figure}
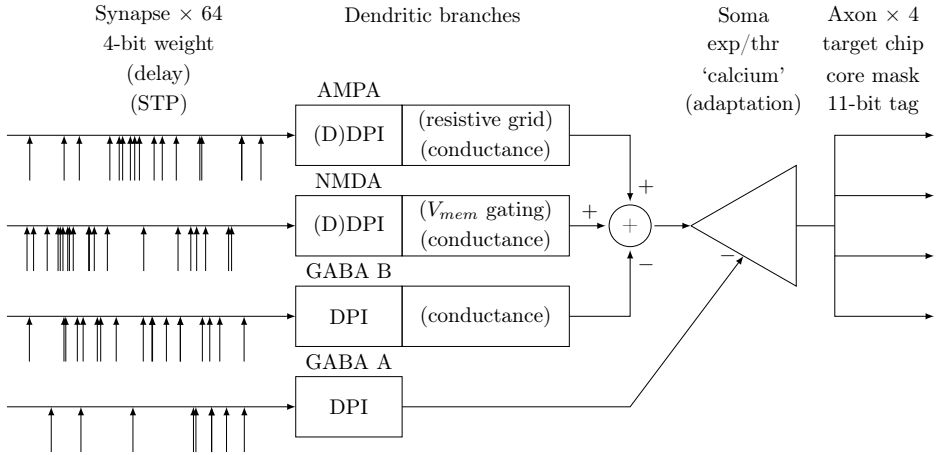

As summarized in Fig.~\ref{fig:neuron-model}, each neuron circuit is composed of synaptic, dendritic and somatic compartments with many conditional blocks for dynamic features, which are constructed in a highly modular way, meaning that all of them can be bypassed with digital latches when not needed. The default state of these latches after reset is always disabled, so the users do not have to disable them by setting parameters to extreme values as in the previous generation.

In order to better monitor and debug the network, the user can select one neuron per core to monitor, the membrane potential of which is directly buffered to an external pin, and multiple other intermediate analog current signals are converted into pulse-frequency modulated signals using spiking analog-to-digital converters (sADC).
In addition, a delay pulse internal to a couple of specific synapses and the homeostasis direction of the monitored neuron are also buffered to external pins. Section \ref{sec:direct-monitoring}-\ref{sec:on-chip-monitoring} include more details about the monitoring.

\begin{figure}
    \centering
    \includegraphics[trim= 140 520 400 140, clip,width=.6\linewidth]{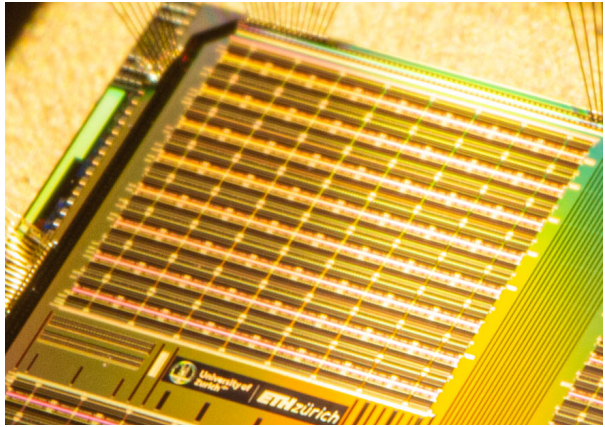}
    \caption{One neural core with 256 neurons in a 16 $\times$ 16 array.}
    \label{fig:one-neural-core}
\end{figure}

\subsection{Specifications}\label{sec:the-dynap-se2-chip}

The specifications of \chipname{} are summarized in Table \ref{tab:specifications}.
\begin{table}[!htbp]
    \caption{Summary of enhanced and new features of \chipname{} compared to a current multi-purpose mixed signal prototyping platform DYNAP-SE \cite{Moradi_etal18} offered by the neuromophic engineering community.}
    \centering
    \small
    \resizebox{\textwidth}{!}{
    \begin{tabular}{c|cc}
        \hline
         & Enhanced features & Novel features \\ \hline
        \multirow{3}*{Resource} & 4 cores with 250 configurable biases. & 8-channel delta-modulated AFE 
        \cite{Sharifshazileh_etal21}. \\
         & 1024 integrate-and-fire neurons. & DVS interface with pre-processing 
         \cite{Richter_etal23}. \\
         & Can target up to $\pm 7 \times \pm 7$ surrounding chips. & 64-channel sADC on-chip monitoring \cite{Corradi_Indiveri15}.\\ \hline
        \multirow{3}*{Neuron} & Exponential \cite{Indiveri_etal11} \& thresholded soma model. & Emulation of calcium current \cite{Bartolozzi_Indiveri07a}. \\
         & Spike-frequency adaptation \cite{Millner_etal10}. & Homeostasis using gain regulation \cite{Qiao_etal17}. \\
         & Monitoring of membrane potential. & Internal state probing with sADC \cite{Corradi_Indiveri15}. \\ \hline
        \multirow{3}*{Synapse} & 64 synapses per neuron. & Quadruple (256) fan-in mode. \\
         & 11-bit content-addressable memory (CAM). & 2-bit precise and mismatched delays \cite{Nielsen_etal17}. \\
         & 4-bit flexibly configurable weight. & Short term plasticity (depression) \cite{Liu_etal14}. \\ \hline
        \multirow{3}*{Dendrite} & Excitatory: AMPA, NMDA (distal). & Alpha function excitatory PSC \cite{Sumislawska_etal16}. \\
         & Inhibitory: GABA\textsubscript{A} (proximal), GABA\textsubscript{B} (distal). & Conductance on distal dendrites \cite{Sumislawska_etal16}. \\
         & Membrane potential-gated NMDA mechanism \cite{Bartolozzi_Indiveri07a}. & 2D resistive grid on AMPA \cite{Benjamin_etal14}. \\ \hline
    \end{tabular}
    }
    \label{tab:specifications}
\end{table}

\section{Common neuromorphic building blocks} \label{sec:common-neuromorphic-building-blocks}


\subsection{Differential pair integrator (DPI)} \label{sec:differential-pair-integrator}




The DPI is current-mode a low-pass filter that enables a wide range of dynamic features in neuromorphic aIC design~\cite{Chicca_etal14}.
It has many advantages such as small area, high power efficiency and good controllability, and is thus used in silicon synapse and neuron designs, as well as longer time constant adaptation~\cite{Millner_etal10} and homeostasis circuits \cite{Qiao_etal17}.
When used as a linear integrator, it can exploit the super-position principle and receive high-frequency spike trains to produce an output that represents the sum of many synapses receiving low-frequency inputs.

\subsubsection{Circuit}

The basic circuit and block diagram of a DPI is shown in Fig.~\ref{fig:dpi_circuit}.

\tikzset{
    DPI/.pic ={
        \ctikzset{multipoles/thickness=3}
        \ctikzset{multipoles/dipchip/width=3}
        \node[dipchip, num pins=6, draw only pins={1-4, 6}, hide numbers, no topmark](C){\tikzpictext};
        \node [right] at (C.bpin 1) {$I_{in}$};
        \node [right] at (C.bpin 2) {$I_{gain}$};
        \node [right] at (C.bpin 3) {$I_{tau}$};
        \node [left] at (C.bpin 4) {$I_{out}$};
        \node [left] at (C.bpin 6) {$V_{out}$};
        \coordinate (-Iin) at (C.pin 1);
        \coordinate (-Igain) at (C.pin 2);
        \coordinate (-Itau) at (C.pin 3);
        \coordinate (-Iout) at (C.pin 4);
        \coordinate (-Vout) at (C.pin 6);
    }
}

\begin{figure}[!htbp]
    \centering
    \scalebox{0.8}{
    \begin{circuitikz}
        \draw (-4, 0) node[pmos] (Qthr) {};
        \draw (Qthr.S) -- ++(1, 0) node[pmos, xscale=-1, anchor=S] (Qc) {};
        \draw (Qc.G) |- (Qc.D) to[short, *-] ++ (0, -0.5) node[nmos, anchor=D] (Qtau) {};
        \draw ($(Qthr.S)!0.5!(Qc.S)$) to[current source, *-, f<=$I_{in}$] ++(0, 2) node[rground, yscale=-1] {};
        \draw (Qc.G) to[short, *-] ++(1, 0) node[nmos, anchor=G] (Qout) {};
        \draw let \p{A}=(Qtau.S) in ($(Qc.G)!0.5!(Qout.G)$) node[anchor=south] {$V_{out}$} to[C=$C$, *-] ++ (0, \y{A}) node[ground] {};
        \draw (Qthr.D) node[ground] {};
        \draw (Qtau.S) node[ground] {};
        \draw (Qout.S) node[ground] {};
        \draw (Qout.D) node[currarrow, rotate=-90] {} -- ++(0, 0.5) node[anchor=south] {$I_{out}$};
        \draw (Qthr.G) node[anchor=east] {$V_{gain}$};
        \draw (Qtau.G) node[anchor=east] {$V_{tau}$};
        
        \draw (4, 0) node[nmos] (Qthr) {};
        \draw (Qthr.S) -- ++(1, 0) node[nmos, xscale=-1, anchor=S] (Qc) {};
        \draw (Qc.G) |- (Qc.D) to[short, *-] ++ (0, 0.5) node[pmos, anchor=D] (Qtau) {};
        \draw ($(Qthr.S)!0.5!(Qc.S)$) to[current source, *-, f>=$I_{in}$] ++(0, -2) node[ground] {};
        \draw (Qc.G) to[short, *-] ++(1, 0) node[pmos, anchor=G] (Qout) {};
        \draw let \p{A}=(Qtau.S) in ($(Qc.G)!0.5!(Qout.G)$) node[anchor=north] {$V_{out}$} to[C=$C$, *-] ++ (0, \y{A}) node[rground, yscale=-1] {};
        \draw (Qthr.D) node[rground, yscale=-1] {};
        \draw (Qtau.S) node[rground, yscale=-1] {};
        \draw (Qout.S) node[rground, yscale=-1] {};
        \draw (Qout.D) node[currarrow, rotate=-90] {} -- ++(0, -0.5) node[anchor=north] {$I_{out}$};
        \draw (Qthr.G) node[anchor=east] {$V_{gain}$};
        \draw (Qtau.G) node[anchor=east] {$V_{tau}$};
        
        \pic[pic text=DPI\textsubscript{N}] (C) at(-2.5, -4.5) {DPI};
        \draw (C-Iin) node[currarrow] {} -- ++(-0.5, 0);
        \draw (C-Igain) -- ++(-0.5, 0);
        \draw (C-Itau) -- ++(-0.5, 0);
        \draw (C-Iout) node[currarrow, xscale=-1] {} -- ++(0.5, 0);
        \draw (C-Vout) -- ++(0.5, 0);
        
        \pic[pic text=DPI\textsubscript{P}] (C) at(5.5, -4.5) {DPI};
        \draw (C-Iin) node[currarrow, xscale=-1] {} -- ++(-0.5, 0);
        \draw (C-Igain) -- ++(-0.5, 0);
        \draw (C-Itau) -- ++(-0.5, 0);
        \draw (C-Iout) node[currarrow] {} -- ++(0.5, 0);
        \draw (C-Vout) -- ++(0.5, 0);
    \end{circuitikz}}
    \caption{N- and P-type DPI circuits and corresponding block diagrams. The output current $I_{out}$ can be thought of as a low-pass filtered version of the input current $I_{in}$. The circuit is designed in current mode, where $I_{x}$ ($x \in \{tau, gain, out\}$) is the current flowing in the diode-connected transistor with voltage $V_x$ of the corresponding type (for example $I_{out}$ and $V_{out}$ in the schematics).}
    \label{fig:dpi_circuit}
\end{figure}
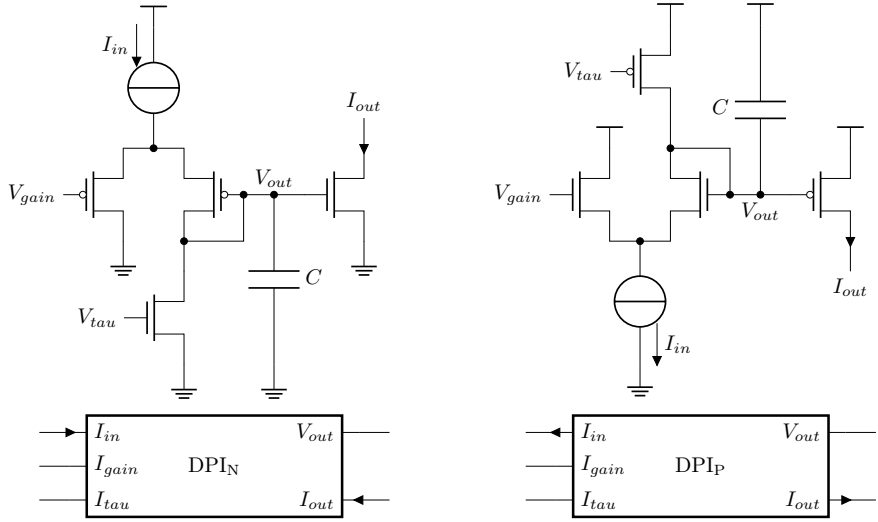

\subsubsection{Equations and typical operating regimes} The most general equation in current mode for the output $I_{out}$ is
\begin{equation}
    \tau \dot{I}_{out} + I_{out} = \frac{I_{in}}{I_{tau}} \frac{I_{gain} I_{out}}{I_{gain} + I_{out}}
\end{equation}
where the time constant $\tau = \frac{C U_T}{\kappa I_{tau}}$. The non-linear equation can be simplified in the three typical operating regimes:
\begin{enumerate}
    \item $I_{out} \gg I_{gain}$,
        \begin{equation} \label{eq:dpi-low-gain}
            \tau \dot{I}_{out} + I_{out} = \frac{I_{gain}}{I_{tau}} I_{in}
        \end{equation}
        which is a first-order linear system with input $I_{in}$ and state variable $I_m$,
    \item $I_{gain} \gg I_{out}$,
        \begin{equation} \label{eq:dpi-high-gain}
            C \dot{V}_{out} = I_{in} - I_{tau}
        \end{equation}
        which is a linear integration of inputs on the membrane capacitor,
    \item $I_{in} \ll I_{tau}$,
        \begin{equation} \label{eq:dpi-leak}
            \tau \dot{I}_{out} + I_{out} = 0 \Leftrightarrow C \dot{V}_{out} = I_{tau}
        \end{equation}
        which is an exponential decay for $I_{out}$ and linear ramp-down for $V_{out}$. 
\end{enumerate}

\subsection{Mirrored output} \label{se:mirrored-output}

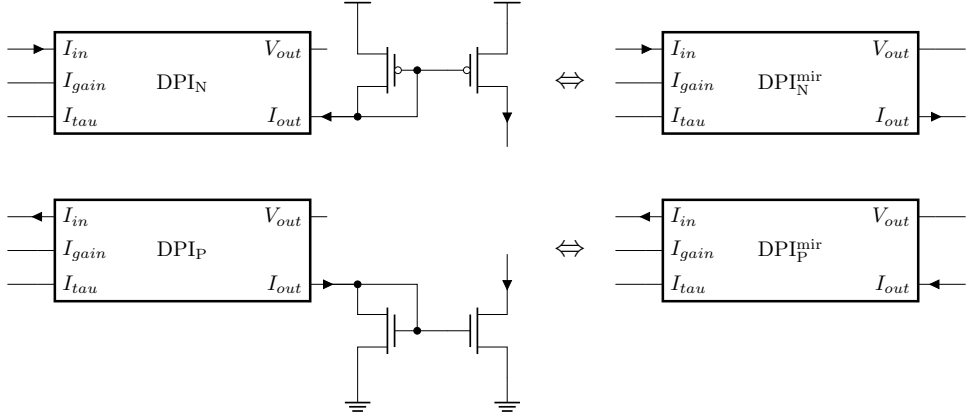
\begin{figure}[!htbp]
    \centering
    \scalebox{0.8}{
    \begin{circuitikz}
        \node at (0, 1.25) {};
        \pic[pic text=$\mathrm{DPI_{N}}$] (C1) at(0, 0) {DPI};
        \draw (C1-Iin) node[currarrow] {} -- ++(-0.5, 0);
        \draw (C1-Igain) -- ++(-0.5, 0);
        \draw (C1-Itau) -- ++(-0.5, 0);
        \draw (C1-Iout) node[currarrow, xscale=-1] {} -- ++(0.5, 0);
        \draw (C1-Iout) to[short, -*] ++(0.5, 0) node[pmos, xscale=-1, anchor=D] (Q1) {};
        \draw (Q1.D) -| (Q1.G) to[short, *-] ++(0.5, 0) node[pmos, anchor=G] (Q2) {};
        \draw (Q1.S) node[rground, yscale=-1] {};
        \draw (Q2.S) node[rground, yscale=-1] {};
        \draw (Q2.D) node[currarrow, rotate=-90] {} -- ++(0, -0.5);
        \pic[pic text=$\mathrm{DPI_{N}^{mir}}$] (C2) at(10, 0) {DPI};
        \draw (C2-Igain) ++(-1.25, 0) node {\Large$\Leftrightarrow$};
        \draw (C2-Iin) node[currarrow] {} -- ++(-0.5, 0);
        \draw (C2-Igain) -- ++(-0.5, 0);
        \draw (C2-Itau) -- ++(-0.5, 0);
        \draw (C2-Iout) node[currarrow] {} -- ++(0.5, 0);
        \draw (C2-Vout) -- ++(0.5, 0);
        \node at (0, -1.25) {};
    \end{circuitikz}}
    \scalebox{0.8}{
    \begin{circuitikz}
        \node at (0, 1.25) {};
        \pic[pic text=$\mathrm{DPI_{P}}$] (C1) at(0, 0) {DPI};
        \draw (C1-Iin) node[currarrow, xscale=-1] {} -- ++(-0.5, 0);
        \draw (C1-Igain) -- ++(-0.5, 0);
        \draw (C1-Itau) -- ++(-0.5, 0);
        \draw (C1-Iout) node[currarrow] {} -- ++(0.5, 0);
        \draw (C1-Iout) to[short, -*] ++(0.5, 0) node[nmos, xscale=-1, anchor=D] (Q1) {};
        \draw (Q1.D) -| (Q1.G) to[short, *-] ++(0.5, 0) node[nmos, anchor=G] (Q2) {};
        \draw (Q1.S) node[ground] {};
        \draw (Q2.S) node[ground] {};
        \draw (Q2.D) node[currarrow, rotate=-90] {} -- ++(0, 0.5);
        \pic[pic text=$\mathrm{DPI_{P}^{mir}}$] (C2) at(10, 0) {DPI};
        \draw (C2-Igain) ++(-1.25, 0) node {\Large$\Leftrightarrow$};
        \draw (C2-Iin) node[currarrow, xscale=-1] {} -- ++(-0.5, 0);
        \draw (C2-Igain) -- ++(-0.5, 0);
        \draw (C2-Itau) -- ++(-0.5, 0);
        \draw (C2-Iout) node[currarrow, xscale=-1] {} -- ++(0.5, 0);
        \draw (C2-Vout) -- ++(0.5, 0);
        \node at (0, -1.25) {};
    \end{circuitikz}}
    \caption{N- and P-type DPI with mirrored output. The new output $I_{out}$ flows in the opposite direction to the original one in Fig.~\ref{fig:dpi_circuit} but has the same magnitude.}
    \label{fig:dpi_circuit_mirrored}
\end{figure}

The output current can be flipped using a current mirror so that it flows in the same direction as the input, as shown in Fig.~\ref{fig:dpi_circuit_mirrored}.

\subsection{Pulse extender} 

As the information in the network is exclusively carried with spikes, which are extremely short duration (sub-nanosecond) digital pulses, they would be largely inconsequential for the analog circuits, thus there must be a way to convert the spikes into analog pulses with a longer duration.
For instance, the input presynaptic spikes have to be converted into analog post-synaptic currents, and the neuron spikes have to trigger refractory periods and negative feedback mechanisms such as spike-frequency adaptation and homeostasis, etc. This conversion is achieved with a class of pulse extender circuits.

\subsubsection{Basic pulse extender} \label{sec:basic-pulse-extender}

The most simple and low power pulse extender circuit is shown in Fig.~\ref{fig:sch_px}.
The pulse width $T_{pulse}$ is controlled by the discharging current $I_{pw}$ in an inversely proportional manner:
\begin{equation} \label{eq:t_pulse}
    T_{pulse} \propto \frac{1}{I_{pw}}
\end{equation}

\tikzset{
    PX/.pic ={
        \ctikzset{multipoles/thickness=3}
        \ctikzset{multipoles/dipchip/width=2.5}
        \node[dipchip, num pins=6, draw only pins={1, 3, 5}, hide numbers, no topmark](C){\tikzpictext};
        \node [right] at (C.bpin 1) {$\overline{\mathrm{event}}$};
        \node [right] at (C.bpin 3) {$I_{pw}$};
        \node [left] at (C.bpin 5) {pulse};
        \coordinate (-event) at (C.pin 1);
        \coordinate (-Ipw) at (C.pin 3);
        \coordinate (-pulse) at (C.pin 5);
    }
}

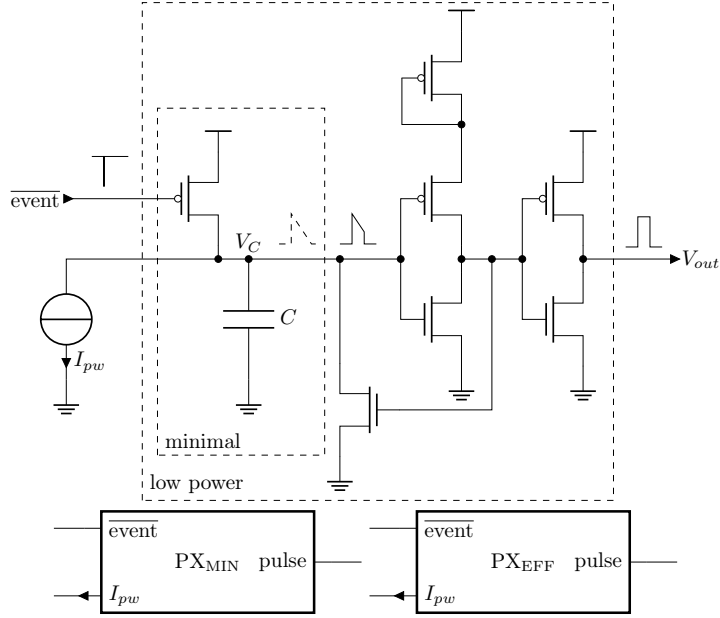
\begin{figure}
    \centering
    \scalebox{0.8}{
    \begin{circuitikz}
        \draw (1, 0) -- (-1.5, 0) to[current source=$I_{pw}$] ++ (0, -2) node[ground] {};
        \draw (1, 1) node[pmos](Qevent) {};
        \draw (Qevent.S) node[rground,yscale=-1] {};
        \draw (Qevent.D) -- (1, 0) to[short, *-*] (4, 0);
        \draw (1.5, 0) node[anchor=south] {$V_{C}$} to[C=$C$, *-] ++(0, -2) node[ground] {};
        \draw[dashed] (2, 0.25) -| ++(0.2, 0.5) -- ++(0.333, -0.5) -- ++(0.167, 0);
        \draw (3, 0.25) -| ++(0.2, 0.5) -- ++(0.2, -0.3) |- ++(0.2, -0.2);
        \draw (3, -2.5) node[nmos, xscale=-1] (Qdischarge) {};
        \draw (5, -1) node[nmos](Qn1) {};
        \draw (5, 1) node[pmos](Qp1) {};
        \draw (5, 3) node[pmos](Qp2) {};
        \draw (Qn1.D) -- (Qp1.D);
        \draw (Qp2.G) |- (Qp2.D) to [short, *-] (Qp1.S);
        \draw (Qp2.S) node[rground, yscale=-1] {};
        \draw (7, -1) node[nmos](Qinvn) {};
        \draw (7, 1) node[pmos](Qinvp) {};
        \draw (Qn1.G) -- (4, -1) -- (4, 1) -- (Qp1.G);
        \draw (Qn1.S) node[ground] {};
        \draw (Qdischarge.S) node[ground] {};
        \draw (Qdischarge.D) to[short, -*] (3, 0);
        \draw (5, 0) to[short, *-*] (5.5, 0) |- (Qdischarge.G);
        \draw (5.5, 0) to[short, -*] (6, 0);
        \draw (Qinvn.G) -- (6, -1) -- (6, 1) -- (Qinvp.G);
        \draw (Qinvp.S) node[rground, yscale=-1] {};
        \draw (Qinvn.S) node[ground] {};
        \draw (Qinvn.D) -- (Qinvp.D);
        
        \draw (Qevent.G) -- ++(-1.5, 0) node[currarrow] {} node[anchor=east] {$\overline{\mathrm{event}}$};
        \draw (Qevent.G) ++ (-0.5, 0.7) -| ++ (-0.4, -0.5) |- ++ (-0.2, 0.5);
        \draw (7, 0) to[short, *-] ++(1.5, 0) node[currarrow] {} node[anchor=west] {$V_{out}$};
        \draw (7, 0) ++ (0.7, 0.2) -| ++ (0.2, 0.5) -- ++ (0.2, 0) |- ++ (0.2, -0.5);
        
        \draw[dashed] (-0.25, -4) node[anchor=south west] {low power} rectangle (7.5, 4.25);
        \draw[dashed] (0, -3.25) node[anchor=south west] {minimal} rectangle (2.75, 2.5);
        
        \draw (0, -4) node {};
        
    \end{circuitikz}}
    \scalebox{0.8}{
    \begin{circuitikz}
        \pic[pic text=PX\textsubscript{MIN}] (C) at(0, 0) {PX};
        \draw (C-Ipw) node[currarrow, xscale=-1] {} -- ++(-0.5, 0);
        \draw (C-event) -- ++(-0.5, 0);
        \draw (C-pulse) -- ++(0.5, 0);
    \end{circuitikz}
    \begin{circuitikz}
        \pic[pic text=PX\textsubscript{EFF}] (C) at(0, 0) {PX};
        \draw (C-Ipw) node[currarrow, xscale=-1] {} -- ++(-0.5, 0);
        \draw (C-event) -- ++(-0.5, 0);
        \draw (C-pulse) -- ++(0.5, 0);
    \end{circuitikz}}
    \caption{Minimal and low power pulse extender. When the active-low input event arrives, the capacitor $C$ immediately charges to $\mathrm{V_{dd}}$, then discharges with current $I_{pw}$. For the minimal pulse extender $\mathrm{PX_{MIN}}$ with only one transistor and one capacitor, the voltage $V_C$ on the capacitor is the output. This circuit is simple, but the output is not clean (dashed waveform) and consumes more power as it stays around $\mathrm{V_{dd}}/2$ longer. For the low-power pulse extender $\mathrm{PX_{EFF}}$, once $V_C$ reaches the switching threshold around $\mathrm{V_{dd}}/4$, positive feedback will discharge the capacitor rapidly (solid line), so the output pulse is cleaner and consumes less power. The switching threshold is shifted down to $\sim\mathrm{V_{dd}}/4$ by the unsymmetrical starved inverter as well as sizing the P-FET physically the same size as the N-FET and resulting in a beneficial pull-up/pull-down drive strength imbalance. With this the capacitance can be significantly smaller while still achieving the same time constant.}
    \label{fig:sch_px}
\end{figure}


\subsubsection{Delayed pulse extender} \label{sec:delayed-pulse-extender}

The pulse extender circuit in Fig.~\ref{fig:sch_px} charges the capacitor immediately to $\mathrm{V_{dd}}$ when the input event arrives, which makes the output pulse also immediate.
If the charging current of the input current is also restricted with an analog parameter, the output pulse will be delayed with reference to the input \cite{Nielsen_etal17}.
The circuit is shown in Fig.~\ref{fig:delayed_pulse_extender}. The delay time $T_{delay}$ is controlled by the charging current $I_{delay}$, and pulse width $T_{pulse}$ by the discharging current $I_{pw}$, both in an inversely proportional manner:
\begin{equation}
    T_{delay} \propto \frac{1}{I_{delay}}, T_{pulse} \propto \frac{1}{I_{pw}}
\end{equation}

\tikzset{
    PXdly/.pic ={
        \ctikzset{multipoles/thickness=3}
        \ctikzset{multipoles/dipchip/width=2.5}
        \node[dipchip, num pins=6, draw only pins={1-3, 5}, hide numbers, no topmark](C){\tikzpictext};
        \node [right] at (C.bpin 1) {$\overline{\mathrm{event}}$};
        \node [right] at (C.bpin 2) {$I_{delay}$};
        \node [right] at (C.bpin 3) {$I_{pw}$};
        \node [left] at (C.bpin 5) {$\overline{\mathrm{pulse}}$};
        \coordinate (-event) at (C.pin 1);
        \coordinate (-Idelay) at (C.pin 2);
        \coordinate (-Ipw) at (C.pin 3);
        \coordinate (-pulse) at (C.pin 5);
    }
}

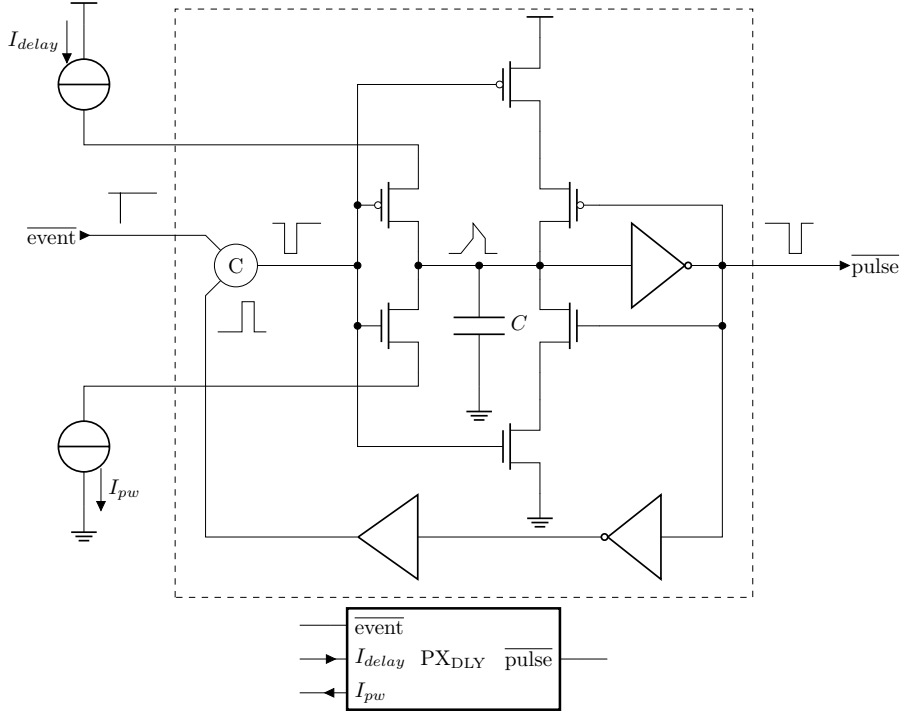
\begin{figure}[!htbp]
    \centering
    \scalebox{.8}{
    \begin{circuitikz}
        \draw (0,0) circle (0.354) node (C) {C};
        \draw (-0.25, 0.25) -- ++ (-0.25, 0.25) -- ++ (-2, 0) node[currarrow] {} node[anchor=east] {$\overline{\mathrm{event}}$};
        \draw (-2.1, 1.2) -| ++(0.2, -0.5) |- ++(0.6, 0.5);
        \draw (3, -1) node[nmos] (Qcn1) {};
        \draw (3, 1) node[pmos] (Qcp1) {};
        \draw (4, 0) to[C=$C$, *-] ++ (0, -2) node[ground] {};
        \draw (5, -3) node[nmos] (Qcn2) {};
        \draw (5, 3) node[pmos] (Qcp2) {};
        \draw (5, -1) node[nmos, xscale=-1] (Qfbn) {};
        \draw (5, 1) node[pmos, xscale=-1] (Qfbp) {};
        \draw (3, 0) to[short, *-*] ++(2, 0) -- ++(1, 0) to[inline not, -*] ++(2, 0);
        \draw (0.354, 0) to[short, -*] (2, 0);
        \draw (Qcp2.G) -| (2, 1) to[short, *-*] (2, -1) |- (Qcn2.G);
        \draw (Qcp1.D) -- (Qcn1.D);
        \draw (Qfbp.D) -- (Qfbn.D);
        \draw (Qcp2.D) -- (Qfbp.S);
        \draw (Qcn2.D) -- (Qfbn.S);
        \draw (Qcp1.S) |- (-2.5, 2) to[current source, f<=$I_{delay}$] ++(0, 2) node[rground, yscale=-1] {};
        \draw (Qcn1.S) |- (-2.5, -2) to[current source, f>=$I_{pw}$] ++(0, -2) node[ground] {};
        \draw (Qcp2.S) node[rground, yscale=-1] {};
        \draw (Qcn2.S) node[ground] {};
        \draw (Qfbp.G) -| (8, -4.5) to[inline not] ++(-3, 0) to[inline buffer] ++(-5, 0) -| (-0.5, -0.5) -- (-0.25, -0.25);
        \draw (-0.3, -1.1) -| ++(0.4, 0.5) -| ++(0.2, -0.5) -- ++(0.2, 0);
        \draw (Qfbn.G) to[short, -*] (8, -1);
        \draw (8, 0) -- ++(2, 0) node[currarrow] {} node[anchor=west] {$\overline{\mathrm{pulse}}$};
        \draw[dashed] (-1, -5.5) rectangle (8.5, 4.25);
        \draw (0.6, 0.7) -| ++(0.2, -0.5) -| ++(0.2, 0.5) -- ++(0.4, 0);
        \draw (3.5, 0.2) -| ++(0.2, 0) -- ++(0.2, 0.25) -- ++(0, 0.25) -- ++(0.2, -0.25) |- ++(0.2, -0.25);
        \draw (8.7, 0.7) -| ++(0.4, -0.5) -| ++(0.2, 0.5) -- ++(0.2, 0);
        \draw (0, -5.5) node {};
    \end{circuitikz}}
    \scalebox{.8}{
    \begin{circuitikz}
        \pic[pic text=PX\textsubscript{DLY}] (C) at(0, 0) {PXdly};
        \draw (C-Idelay) node[currarrow] {} -- ++(-0.5, 0);
        \draw (C-Ipw) node[currarrow, xscale=-1] {} -- ++(-0.5, 0);
        \draw (C-event) -- ++(-0.5, 0);
        \draw (C-pulse) -- ++(0.5, 0);
    \end{circuitikz}}
    \caption{Delayed pulse extender. The C-element \cite{Muller55} (shown as \copyright) is an asynchronous digital circuit that changes its output to $X$ when both inputs are equal to $X$. When the active-low event arrives, if there is no output pulse (1), the output of the C-element goes from 1 to 0, which starts the charging of the capacitor with current $I_{delay}$. When the voltage on the capacitor exceeds the threshold of the inverter, the output pulse becomes active (0) and positive feedback charges the capacitor to $\mathrm{V_{dd}}$ immediately. The output of the C-element then goes to 1, which starts the discharging of the capacitor with current $I_{pw}$. When the voltage on the capacitor drops below the threshold of the inverter, the output pulse finishes (1).}
    \label{fig:delayed_pulse_extender}
\end{figure}

\subsubsection{Loss of information}

Both pulse extension and delay mechanisms will make each spike take longer.
The important edge case is when another event arrives before the pulse of the previous event finishes.
From the circuit and information theoretic perspective, since the `time left' information (for either delay or pulse width) is stored as the voltage on the capacitor, it is impossible to keep track of multiple of them with only one state variable.
If two pulses overlap, one of them must be dropped. Because physical systems are causal, the second pulse cannot remove the already started one but only overwrite the remaining part of it. 

For the low-power pulse extender circuit, the capacitor will be recharged to $\mathrm{V_{dd}}$ immediately when the second event arrives, thus the pulse restarts.
Mathematically, the output pulse is the union (logical OR) of the incoming pulses. 

For the delayed pulse extender circuit, the charging can only start when the output pulse is inactive, otherwise the output of the C-element will remain at 1.
If another input event arrives during the delay phase or in the extreme case during the transition between delay and pulse phase, the output of the C-element will still be 0.
In both cases, the output of the C-element does not change, meaning that the information is dropped. In other words, if the inter-spike interval is shorter than the delay, the second spike will be ignored. 

\subsection{Event low pass filter} \label{sec:low-pass-filter}

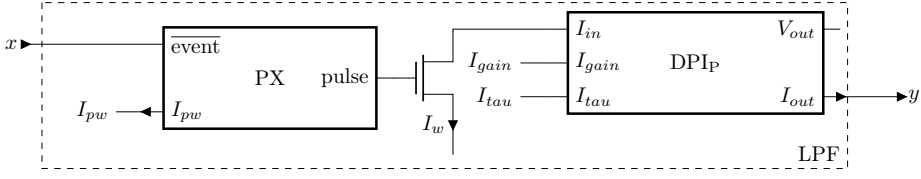
\begin{figure}[!htbp]
    \centering
    \scalebox{.8}{
    \begin{circuitikz}
        \pic[pic text=PX] (PX) at (0, 0) {PX};
        \pic[pic text=DPI\textsubscript{P}] (DPI) at (7, 0.25) {DPI};
        \draw (PX-pulse) node[nmos, anchor=G] (Q) {};
        \draw (Q.D) |- (DPI-Iin);
        \draw (PX-Ipw) node[currarrow, xscale=-1] {} -- ++(-0.5, 0) node[anchor=east] {$I_{pw}$};
        \draw (Q.S) node[currarrow, rotate=-90] {} node[anchor=east] {$I_{w}$} -- ++(0, -0.5);
        \draw (DPI-Igain) -- ++(-0.5, 0) node[anchor=east] {$I_{gain}$};
        \draw (DPI-Itau) -- ++(-0.5, 0) node[anchor=east] {$I_{tau}$};
        \draw (DPI-Iout) node[currarrow] {} -- ++(1, 0) node[currarrow] {} node[anchor=west] {$y$};
        \draw (PX-event) -- ++(-2, 0) node[currarrow] {} node[anchor=east] {$x$};
        \draw[dashed] (-3.75, 1.25) rectangle (9.5, -1.5) node[anchor=south east] {LPF};
    \end{circuitikz}}
    \caption{Event low pass filter consisting of a pulse extender PX and a DPI. The input $x$ is a set of discrete events (treated as sum of Dirac functions) and the output $y$ is an analog current waveform.}
    \label{fig:low-pass-filter}
\end{figure}

When a pulse extender is combined with a DPI as shown in Fig.~\ref{fig:low-pass-filter}, it serves as an event low-pass filter (LPF).

Let the (active low) input be $x(t) = \sum_{i =1}^{N} \delta \left(t - t_i\right)$ where the $t_i$ are the spike times, the pulse width of the pulse extender be $T_{pulse}$, the time constant and threshold parameters for the DPI be $I_{tau} and I_{gain}$, and the weight parameter be $I_w$. $\tau=\frac{CU_T}{\kappa I_{tau}}$ and
\begin{equation}
    W = \frac{I_{gain} I_w}{I_{tau}} \frac{T_{pulse}}{\tau} = \frac{\kappa I_{gain} I_w T_{pulse}}{C U_T} \propto I_{gain} I_w T_{pulse}
\end{equation}

If $\forall i = 1, \cdots, N, t_{i+1} - t_i > T_{pulse}$ and $\frac{I_{tau}}{I_{w}} \tau \ll T_{pulse} \ll \tau$, the combined circuit is a first-order low-pass filter with transfer function
\begin{equation}
    \frac{Y(s)}{X(s)} = \frac{\tau W}{\tau s + 1}
\end{equation}

Similarly for the delayed pulse extender with the extra delay parameter $T_{delay}$, if $\forall i = 1, \cdots, N, t_{i+1} - t_i > T_{delay} + T_{pulse}$ and $\frac{I_{tau}}{I_{w}} \tau \ll T_{pulse} \ll \tau$, the combined circuit is a delayed first-order low-pass filter with transfer function
\begin{equation}
    \frac{Y(s)}{X(s)} = \frac{\tau W e^{-T_{delay} s}}{\tau s + 1}
\end{equation}

If we plug in $X(s) = \mathcal{L} \left[x(t)\right] = \sum_{i = 1}^{N} e^{-t_i s}$, the integral
\begin{equation}
    \int_0^\infty y(t) \mathrm{d}t = \lim_{s \rightarrow 0} Y(s) = \tau W N
\end{equation}
which implies that the system is linear, and the total output charge per input event is
\begin{equation}
    Q = \tau W = \frac{I_{gain} I_w}{I_{tau}} T_{pulse}
\end{equation}

The output only depends on hyperparameters $\tau$, $W$ and $T_{delay}$ or even $Q$ and $T_{delay}$ if $\tau \ll t_{i+1} - t_i~(i = 1, \cdots, N)$.

\subsection{Digital-to-analog converters}

The parameters required to properly operate the analog circuits in the chip are generated on chip by on-chip programmable digital-to-analog converters (DAC).

Because of the large scale of the neural network, i.e.~1024 neurons $\times~(\sim$20 somatic parameters $+\sim$20 parameters for the four dendrites $+$ 64 synapses per neuron $\times$ 14 synaptic parameters$)$, if every neuron and every synapse would have individually configurable parameters, there would be around one million parameters to set.
As a trade-off, the neurons are divided into four cores of 256 neurons each, and most of the parameters are shared across all neurons and synapses within a core, and implemented with separate parameter generator DACs for each core.
A very few but important cases such as the individual synaptic delays and weights are implemented with a flexible DAC mechanism, consisting of several global parameters for the `base' currents and individual digital latches in each individual unit to chose a binary combination of the `base' currents.

\subsection{Parameter generator} \label{parameter-generator}

\begin{figure}
    \centering
    \includegraphics[trim= 250 220 580 250, clip,width=.6\linewidth]{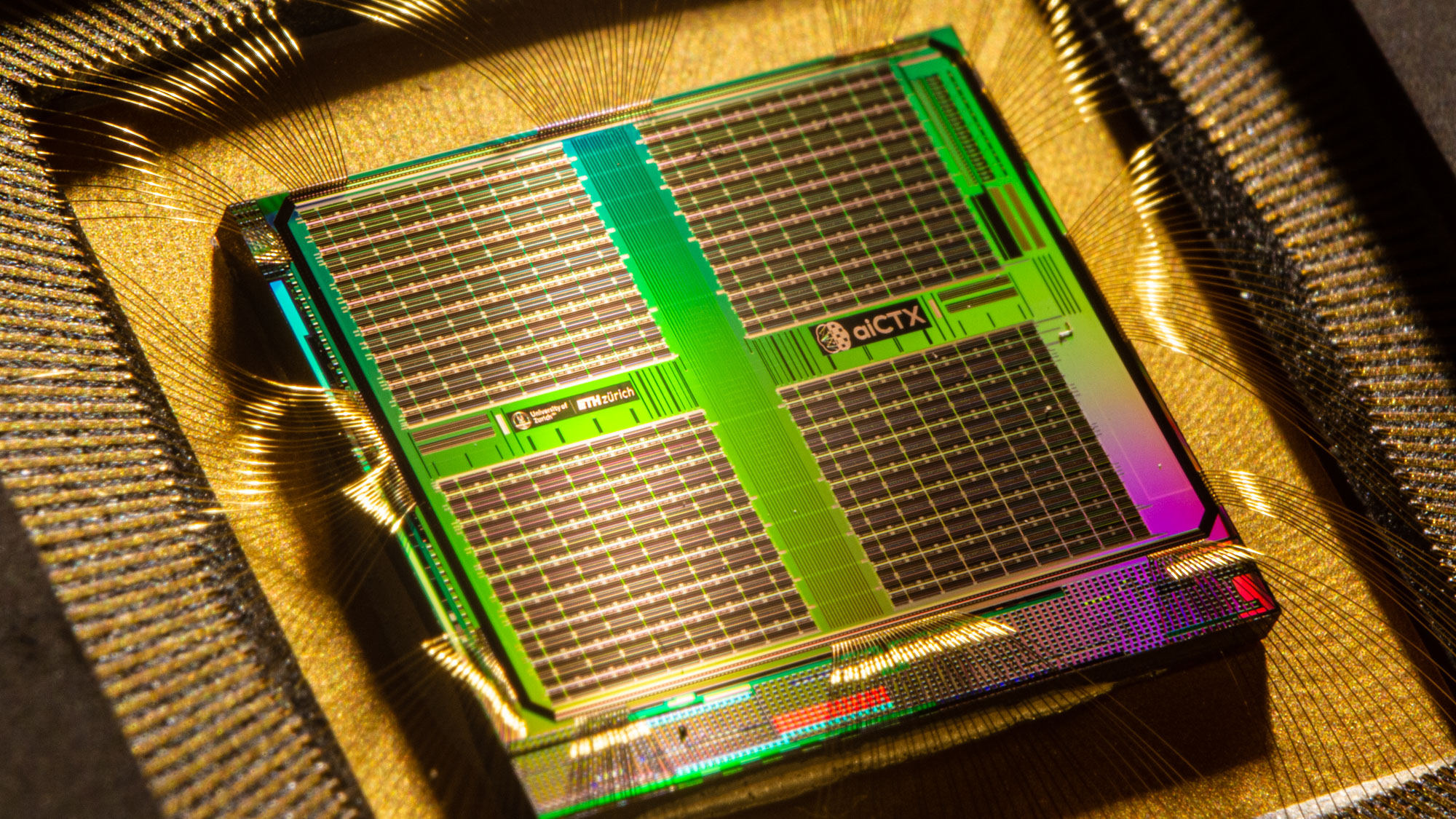}
    \caption{One DAC (two horizontal structures) with the adjacent sADC block (bright rectangle) between two adjacent neural cores.}
    \label{fig:one-dac}
\end{figure}

The current-based parameter generator used in this chip 
generates accurate analog currents over a very large dynamic range~\cite{Yang_etal12}.
This parameter generator is enhanced with a proportional to absolute temperature (PTAT) and complementary to absolute temperature (CTAT) current reference for current temperature stabilization.
The general formula for any current parameter $I_{\mathrm{parameter}}$ is
\begin{equation} \label{eq:pg}
    I_{\mathrm{parameter}} = k_{\mathrm{parameter}} I_{\mathrm{coarse}}(n_{\mathrm{coarse}}) \frac{\mathrm{n_{fine}}}{255}
\end{equation}
where integers $n_{\mathrm{coarse}} \in [0, 5]$, $n_{\mathrm{fine}} \in [0, 255]$. $k_{\mathrm{parameter}}$ is a scaling factor which is roughly constant for all $n_{\mathrm{coarse}}$ and $n_{\mathrm{fine}}$ values, but a more precise non-ideality correction from simulations based on the transistor type and size is also available.

Estimates of the values of the `base' currents $I_{\mathrm{coarse}}(n_{\mathrm{coarse}})$ are shown in Table \ref{tab:i-coarse}.

\begin{table}[!htbp]
    \centering
    \caption{Nominal $I_{\mathrm{coarse}}$ value for each $n_{\mathrm{coarse}}$ value}
    \begin{tabular}{|c|c|c|c|c|c|c|c|}
        \hline
        $n_{\mathrm{coarse}}$ & 0 & 1 & 2 & 3 & 4 & 5 \\
        \hline
        $I_{\mathrm{coarse}}$ & 70 pA & 550 pA & 4.45 nA & 35 nA & 0.28 $\mu$A & 2.25 $\mu$A \\
        \hline
    \end{tabular}
    \label{tab:i-coarse}
\end{table}

It is important to note that the error in these estimates increases with the coarse value, and because of mismatch, different $(n_{\mathrm{coarse}},n_{\mathrm{fine}})$ combinations that produce the same result according to Eq.~(\ref{eq:pg}) may give different results on actual hardware.
In the case of very low currents, $n_{\mathrm{coarse}} = 0$ always gives the highest accuracy.
Therefore, it is recommended to  always use lower $n_{\mathrm{coarse}}$ values when possible.
Especially when $n_{\mathrm{fine}} = 0$, the parameter generator outputs the dark current of the corresponding transistor, and can be very different for different $n_{\mathrm{coarse}}$ values.
As for other implementations~\cite{Yang_etal12}, a small-scale non-monotonicity, caused by a large transistor stack moving out of saturation in the current branch also exists in this implementation and can be corrected via calibration with a pre-recorded look-up table.
For very small currents the DAC requires a settling time for the parameters to reach their steady-state programmed values, which can take up to several seconds.

For circuit parameters that are in the voltage-domain instead of the current one, the voltage $V_{\mathrm{parameter}}$  at the gate of the diode-connected transistor of the appropriate type that conducts the parameter current in sub-threshold is given by
\begin{equation}
    V_{\mathrm{parameter}} = 
    \left\{
    \begin{array}{rl}
        \frac{U_T}{\kappa} \ln \frac{I_{\mathrm{parameter}}}{I_0} & (\mathrm{if~N-type}) \\
        \mathrm{V_{dd}} - \frac{U_T}{\kappa} \ln \frac{I_{\mathrm{parameter}}}{I_0} & (\mathrm{if~P-type})
    \end{array} \right.
\end{equation}

\subsection{Flexible DAC} \label{sec:flexible-dac}

\begin{figure}[!htbp]
    \centering
    \scalebox{.8}{
    \begin{circuitikz}
        \draw[dashed] (0, 0) node[pmos] (Q0) {};
        \draw (Q0.G) node[anchor=east] {$\overline{x_0}$};
        \draw (Q0.S) to[current source, f<=$I_0$] ++ (0, 2) node[rground, yscale=-1] {};
        \draw (2, 0) node[pmos] (Q1) {};
        \draw (Q1.G) node[anchor=east] {$\overline{x_1}$};
        \draw (Q1.S) to[current source, f<=$I_1$] ++ (0, 2) node[rground, yscale=-1] {};
        \node at (3, 0) {\Large{$\cdots$}};
        \draw (5, 0) node[pmos] (Q2) {};
        \draw (Q2.G) node[anchor=east] {$\overline{x_n}$};
        \draw (Q2.S) to[current source, f<=$I_n$] ++ (0, 2) node[rground, yscale=-1] {};
        \draw (Q0.D) -- (0, -1) -| (Q1.D);
        \draw (Q2.D) |- (2.5, -1);
        \draw (2, -1) to[short, *-*] (2.5, -1) -- ++(0, -0.5) node[currarrow, rotate=-90] {} node[anchor=west] {$I_{out}$} -- ++ (0, -0.5);
    \end{circuitikz}}
    \caption{Flexible DAC of $n+1$ bits with minimal current (including $\overline{x_0}$ transistor, dashed line) and $n$ bits without it (dashed transistor connected to $\overline{x_0}$ bypassed).}
    \label{fig:flexible-dac}
\end{figure}
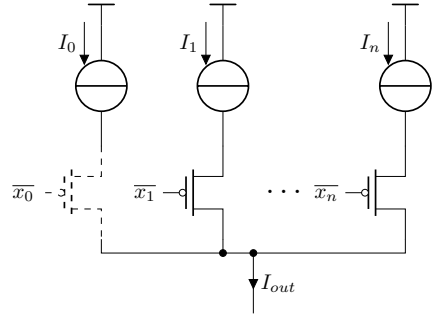

For the 4-bit synaptic weight and 2-bit delay, in order to achieve maximum flexibility, a customized DAC is used. The circuit (in P-type) is shown in Fig.~\ref{fig:flexible-dac}.

The base currents $I_{b0}$ through $I_{bn}$ come from the parameter generator, and the digital configurations $x_0$ through $x_n$ are stored in latches, The output current follows
\begin{equation} \label{eq:idac}
    I_{out} = \sum_{i=0}^n x_i I_{bi}
\end{equation}

If an always-on minimal current is wanted, the P-FET connected to $\overline{x_0}$ could be bypassed, which implies $x_0 \equiv 1$ in Eq.~(\ref{eq:idac}).

If we set $I_{i} = 2^{-i} I_{b0}$ (for $i=1, \cdots, n$), the flexible DAC could be used as a normal $n+1$ bit DAC. If higher dynamic range is needed, the different $I_{bi}$'s can also be very different, but the bit resolution will be lower as a trade off.

\section{Silicon neuron circuits} \label{sec:analog-neuronal-circuits}
\subsection{Somatic compartment} \label{sec:somatic-compartment}

The center of the silicon neuron is the integrate-and-fire soma circuit.
Based on the desired `firing' mechanisms, there are two switchable somatic models on the chip:
\begin{itemize}
    \item Thresholded: the neuron fires when the membrane potential reaches a threshold;
    \item Exponential: the neuron receives positive feedback that drives spike output \cite{Brette_Gerstner05}.
\end{itemize}

\begin{figure}[!htbp]
    \centering
    \scalebox{.8}{
    \begin{tikzpicture}
        \draw (0, 3) rectangle (3,0) node[pos=.5] {somatic DPI\textsubscript{N}};
        \draw (1.5, 3.25) node {integrate};
        \draw[-latex] (-1, 2.75) node[left] {leak} -- ++ (1, 0);
        \draw[-latex] (-1, 2.125) node[left] {gain} -- ++ (1, 0);
        \draw[-latex] (-1, 1.5) node[left] {refractory} -- ++ (1, 0);
        \draw[-latex] (-1, 0.875) node[left] {dendritic input} -- ++ (1, 0);
        \draw[-latex] (-1, 0.25) node[left] {somatic input} -- ++ (1, 0);
        \draw[-latex] (3, 1.5) -- node[above] {\begin{tabular}{c} membrane \\ potential \end{tabular}} ++(2, 0) ;
        \draw (7.25, 3.25) node {fire};
        \draw (5, 1.5) -- ++ (0, 0.375) -- ++ (0.5, 0.875) -- ++ (0, -2.5) -- ++ (-0.5, 0.875) -- ++ (0, 0.375);
        \draw[-latex] (-1, -0.375) node[left] {type} -| (5.25, 0.75);
        \draw[-latex] (5.5, 2.25) -- ++ (0.5, 0);
        \draw (6, 1.75) rectangle (8.5, 2.75) node[pos=.5] {thresholded};
        \draw[-latex] (8.5, 2.25) -- ++ (0.5, 0);
        \draw[-latex] (5.5, 0.75) -- ++ (0.5, 0);
        \draw (6, 0.25) rectangle (8.5, 1.25) node[pos=.5] {exponential};
        \draw[-latex] (8.5, 0.75) -- ++ (0.5, 0);
        \draw (9, 1.5) -- ++ (0, 1.25) -- ++ (0.5, -0.875) -- ++ (0, -0.75) -- ++ (-0.5, -0.875) -- ++ (0, 1.25);
        \draw[-latex] (5.25, -0.375) -| (9.25, 0.75);
        \draw[-latex] (9.5, 1.5) -- ++ (0.5, 0) node[right] {spike};
        \draw[-latex] (9.75, 1.5) |- ++ (-0.75, -3.5);
        \draw (9, -2.75) rectangle (6.5, -1.25) node[pos=.5] {\begin{tabular}{c} pulse \\ extender \end{tabular}};
        \draw[-latex] (6.5, -2) -- (6, -2) |- ++ (-0.5, 0.75);
        \draw[-latex] (6, -2) |- ++ (-0.5, -0.75);
        \draw (5.5, -1.75) rectangle (3, -0.75) node[pos=.5] {adaptation};
        \draw[-latex] (3, -1.25) -| (2.375, 0);
        \draw (5.5, -2.25) rectangle (3, -3.25) node[pos=.5] {homeostasis};
        \draw[-latex] (3, -2.75) -| (1.75, 0);
        \draw[-latex] (-1, -1) node[left] {dc} -| (0.5, 0);
        \draw[-latex] (-1, -1.625) node[left] {kill} -| (1.125, 0);
        \draw[dashed] (-2.25, 0) -- (0, 0) -- (5, 1.125) -- (5.5, 1.5) -- (9, 1.5) -- (9.5, 1.125) -- (10, 1.125) -- (10, -3.5) -- (-2.25, -3.5) -- (-2.25, 0) node[pos=.2, right] {\begin{tabular}{c} conditional functions \\ enabled by latches\end{tabular}};
    \end{tikzpicture}}
    \caption{Somatic circuit block diagram. All the conditional functions within the dashed outline can be disabled or bypassed using digital latches.}
    \label{fig:somatic-circuit-block-diagram}
\end{figure}
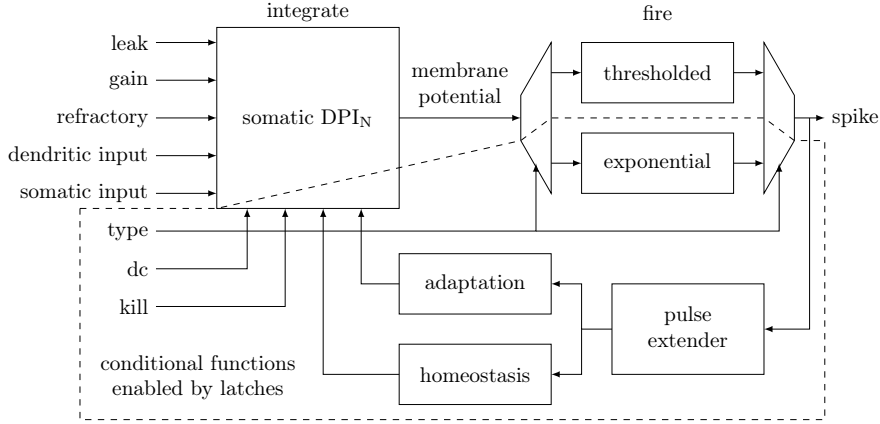

In addition, there are conditional spike-frequency adaptation circuit~\cite{Millner_etal10} and homeostasis circuits~\cite{Qiao_etal17} that can be activated on either model. The overall architecture of the somatic circuit is shown in Fig.~\ref{fig:somatic-circuit-block-diagram}.

\subsubsection{Somatic DPI -- information integration}
\label{sec:soma-dpi}
The integration of information on the soma is achieved with the N-type DPI circuit introduced in Section \ref{sec:differential-pair-integrator}.
There are two basic parameters to control the somatic DPI --  the leak ($\mathsf{SOIF\_LEAK}$ parameter, or the $I_{tau}$ of the DPI) and the gain ($\mathsf{SOIF\_GAIN}$ parameter, or the $I_{gain}$ of the DPI).
The neuron receives post-synaptic current $I_{dendritic}$ from three dendritic branches AMPA, NMDA and GABA\textsubscript{B}, and the somatic current $I_{somatic}$ from shunting inhibitory dendrite GABA\textsubscript{A}. The output is the membrane potential $I_{mem}$ in current mode or $V_{mem}$ in voltage mode.

The most commonly used conditional function is the constant DC current injection (enabled using the latch $\mathsf{SO\_DC}$ and configured with the $\mathsf{SOIF\_DC}$ parameter), which goes into the input branch of the DPI, together with the dendritic input $I_{dendritic}$.
The DC input can be used to set a proper resting potential and even drive a constant firing rate.
One can also turn off any specific neuron  using the latch $\mathsf{SOIF\_KILL}$.

Mathematically, the DPI inputs corresponding to Section \ref{sec:differential-pair-integrator} are
\begin{equation}
    I_{in} = \max \left(I_{dendritic} + I_{DC}, 0\right)
\end{equation}

\begin{equation}
    I_{tau} = I_{leak} + I_{somatic}
\end{equation}

\begin{figure}[!htbp]
    \centering
    \includegraphics[width=0.6\textwidth]{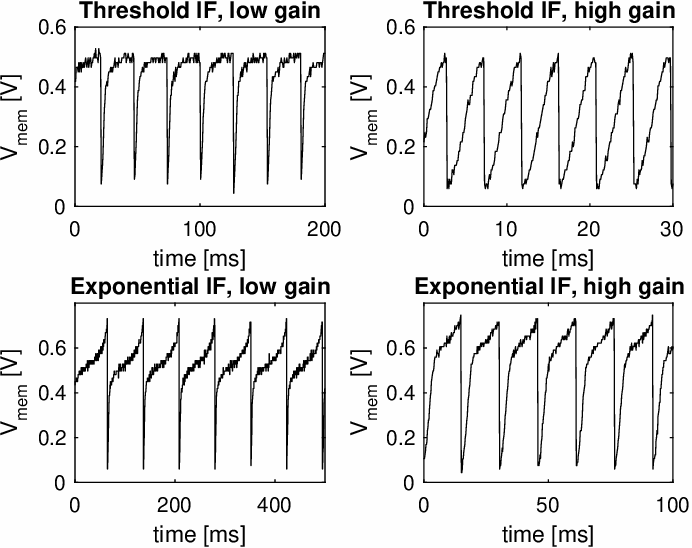}
    \caption{Comparison of the two somatic models in different operating regimes. With a lower gain value (left), the integration phase is logarithmic (linear $I_{mem}$ in Eq.~(\ref{eq:dpi-low-gain})). With a higher gain value (right), the integration phase is linear (exponential $I_{mem}$  in Eq.~(\ref{eq:dpi-low-gain})). The top two plots show the the thresholded model with firing threshold set to around 0.5\,V. The bottom two plots show the exponential model, where $V_{mem}$ has an exponentially increasing shape that leads to the neuron firing. While we show data for the voltage across the output capacitor of the circuits, the neuron uses the current resulting from the voltage across the capacitor. This is given by the exponential of the plotted voltage and it is affected by the relevant transistor variables (e.g., $U_{T}$, $\kappa$)~\cite{Mead89}.}
    \label{fig:comparison-of-two-somatic-models-in-different-operation-regimes}
\end{figure}

Figure \ref{fig:comparison-of-two-somatic-models-in-different-operation-regimes} shows the membrane voltage $V_{mem}$ waveform recorded for a neuron on the chip for the two somatic models with the same DC input but different gains.

\subsubsection{Biologically plausible time constant}

The somatic DPI employs a 7.72\,pF capacitance to achieve a biologically plausible time constant.
When the leak of the neuron is set to its minimum, which is the leakage current of the transistor, the slew rate of $V_{mem}$ can achieve 108 $\pm$ 12 mV/s (measured across one core). Thus a single neuron can hold a `memory' for up to about five seconds, enabling processing of signals on a biologically plausible timescale.


The biologically plausible time constant is a trade off and the reason for a comparatively higher power consumption compared to other systems.
Measurement results and discussion on the power consumption is presented in more detail in subsection \ref{sec:two-models-for-spike-generation}.

\subsubsection{Refractory period -- maximum firing rate}

After the spike is generated, the neuron enters a state in which integration is blocked: this is the (absolute) refractory period in biology.
It is an important computational feature, as it sets an upper limit on the firing rate and introduces non-linearity.
The refractory period circuit as shown in Fig.~\ref{fig:refractory_circuit}.
The length of the refractory period is controlled by the discharging current $I_{\mathrm{refractory}}$ ($\mathsf{SOIF\_REFR}$ parameter).
Based on Eq.~(\ref{eq:t_pulse}), the maximum firing rate $r_{max}$ or equivalently the inverse of the length of the refractory period $T_{\mathrm{refractory}}$ is proportional to the recharging current:
\begin{equation}
    r_{max} = \frac{1}{T_{\mathrm{refractory}}} \propto I_{\mathrm{refractory}}
\end{equation}

\tikzset{
    REFR/.pic ={
        \ctikzset{multipoles/thickness=3}
        \ctikzset{multipoles/dipchip/width=2.5}
        \node[dipchip, num pins=6, draw only pins={1-4, 6}, hide numbers, no topmark](C){\tikzpictext};
        \node [right] at (C.bpin 1) {$\overline{\mathrm{spike}}$};
        \node [right] at (C.bpin 3) {$V_{\mathrm{refractory}}$};
        \node [left] at (C.bpin 4) {$\overline{\mathrm{req}}$};
        \node [right] at (C.bpin 2) {$\overline{\mathrm{ack}}$};
        \node [left] at (C.bpin 6) {refractory};
        \coordinate (-spike) at (C.pin 1);
        \coordinate (-Vrefractory) at (C.pin 3);
        \coordinate (-ack) at (C.pin 2);
        \coordinate (-req) at (C.pin 4);
        \coordinate (-refractory) at (C.pin 6);
    }
}

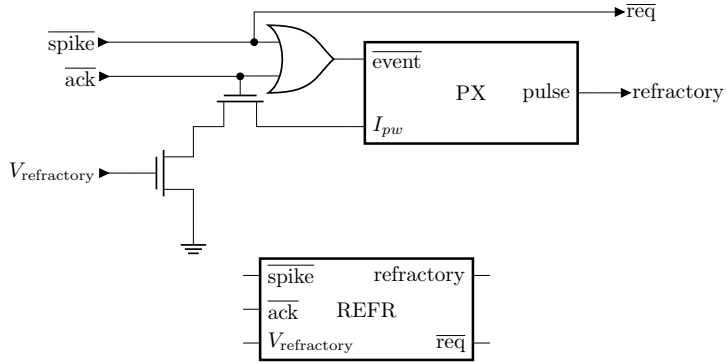
\begin{figure}[!htbp]
    \centering
    \scalebox{.8}{
    \begin{circuitikz}
        \pic [pic text=PX] (PX) at(0, 0) {PX};
        \draw (PX-event) node[or port, anchor=out] (OR) {};
        \draw (PX-Ipw) -- ++(-1, 0) node[nmos, rotate=-90, anchor=D, xscale=0.85] (Qack) {};
        \draw (Qack.S) node[nmos, anchor=D] (Qrefr) {};
        \draw (OR.in 2) -| (Qack.G) to[short, *-] ++(-2.25, 0) node[currarrow] {} node[anchor=east] {$\overline{\mathrm{ack}}$};
        \draw (OR.in 1) to[short, *-] ++ (-2.5, 0) node[currarrow] {} node[anchor=east] {$\overline{\mathrm{spike}}$};
        \draw (OR.in 1) |- ++(6, 0.5) node[currarrow] {} node[anchor=west] {$\overline{\mathrm{req}}$};
        \draw (Qrefr.S) node[ground] {};
        \draw (PX-pulse) -- ++(0.5, 0) node[currarrow] {} node[anchor=west] {refractory};
        \draw (Qrefr.G) -- ++(-0.5, 0) node[currarrow] {} node[anchor=east] {$V_{\mathrm{refractory}}$};
    \end{circuitikz}}
    \scalebox{.8}{
    \begin{circuitikz}
        \pic [pic text=REFR] (C) at(0, 0) {REFR};
    \end{circuitikz}}
    \caption{Refractory circuit and its block diagram. It combines the pulse extender from Section \ref{sec:basic-pulse-extender} with event routing handshaking. In the idle state both the request ($\overline{\mathrm{req}}$) and acknowledge ($\overline{\mathrm{ack}}$) signals are inactive (1). When the neuron emits a spike, ($\overline{\mathrm{spike}} = 0$), $\overline{\mathrm{req}}=0$ is sent to the encoder, which returns $\overline{\mathrm{ack}} = 0$. When both $\overline{\mathrm{req}} = \overline{\mathrm{ack}} = 0$, the pulse extender is triggered, which discharges the neuron until the spike disappears ($\overline{\mathrm{spike}} = \overline{\mathrm{req}} = 1$). The encoder then releases $\overline{\mathrm{ack}}$ ($\overline{\mathrm{ack}} = 1$) and the refractory period starts by discharging at a rate determined by $I_{\mathrm{refractory}}$, during which the neuron's membrane potential is clamped to ground.}
    \label{fig:refractory_circuit}
\end{figure}

The capacitance of the refractory period pulse extender is about 2\,pF.
The longest refractory period is achieved when the parameter is set to its minimum value.
The measurement result across one core shows that the maximum refractory period for the thresholded model is 1.58 $\pm$ 0.10\,s, and 0.748 $\pm$ 0.045\,s for the exponential model.
The difference is because the pulse extender circuits are different in the two models.
The thresholded model uses the low-power pulse extender and the exponential model uses the simplified minimal pulse extender without positive feedback.
The latter also has the problem that there might be multiple events generated for one neuron spike, which makes the exponential model unsuitable for building a complex network.

\subsubsection{Two models for spike generation} \label{sec:two-models-for-spike-generation}

The two leaky integrate-and-fire (I\&F) neuron models are the thresholded I\&F model adapted from \cite{Qiao_Indiveri16} as an intermediate development step towards the later \cite{Rubino_etal20} and the adapted exponential I\&F model as shown in \cite{Indiveri_etal10}.
They share the integration circuit described in subsection \ref{sec:soma-dpi}, but have different ways of generating spikes.
The two models are selected using the $\mathsf{SOIF\_TYPE}$ latch (default 0 $=$ thresholded model, 1 $=$ exponential model).

\begin{enumerate}
    \item Thresholded I\&F model


The thresholded I\&F generates a spike whenever the membrane potential ($I_{mem} = I_{out}$ of the somatic DPI) exceeds a certain threshold $I_{\mathrm{spkthr}}$ (controlled by the parameter $\mathsf{SOIF\_SPKTHR}$).
The circuit is shown in Fig.~\ref{fig:thr_circuit}.
The generated spike will give a positive feedback to the DPI by charging $I_{mem}$ to its maximum immediately, thus the spike pulse width is just the time for the following asynchronous digital encoder to respond and can be as low as a few nanoseconds (thus the ramp-up of $I_{mem}$ is too sharp to be buffered to the monitoring pin and cannot be seen) and it is more power efficient due to being shorter.
The top two plots in Fig.~\ref{fig:comparison-of-two-somatic-models-in-different-operation-regimes} show the firing pattern in the thresholded model.
Measurement results show that it consumes $150 pJ$ per somatic spike when spiking at 80\,Hz for the full soma operation, including the integration of the DC input.

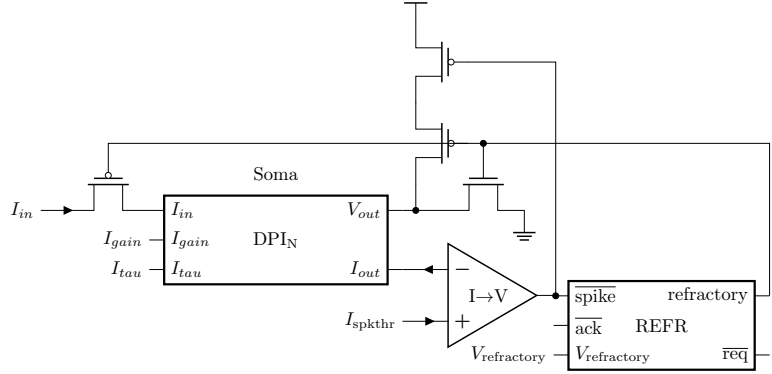
\begin{figure}[!htbp]
    \centering
    \scalebox{.7}{
    \begin{tikzpicture}
        \pic[pic text=DPI\textsubscript{N}] (Soma) at (0, 0) {DPI};
        \draw (0, 1.25) node {Soma};
        \draw (Soma-Iout) -- ++(0.5, 0) node[currarrow, scale=-1] {} node[op amp, anchor=-] (amp) {I$\rightarrow$V};
        \pic [pic text=REFR] (REFR) at(7.25, -1.618) {REFR};
        \draw (Soma-Vout) -| ++(0.25, 0.5) node[pmos, xscale=-1, anchor=D] (Qrefrp) {};
        \draw (Qrefrp.S) node[pmos, xscale=-1, anchor=D] (Qspike) {};
        \draw (amp.out) |- (Qspike.G);
        \draw (Soma-Vout) ++(0.25, 0) to[short, *-] ++(0.5, 0) node[nmos, xscale=-1, rotate=-90, anchor=D] (Qrefrn) {};
        \draw let \p{A}=(Qrefrn.G), \p{B}=(Qrefrp.G) in (\x{A},\y{B}) to[short, *-] (Qrefrn.G);
        \draw let \p{A}=(amp.out), \p{B}=(REFR-spike) in (\x{A},\y{B}) to[short, *-] (REFR-spike);
        \draw (Soma-Iin) node[pmos, anchor=D, rotate=90, xscale=-1] (Qrefri) {};
        \draw (REFR-refractory) |- (Qrefrp.G) -| (Qrefri.G);
        \draw (Qspike.S) node[rground, yscale=-1] {};
        \draw (Qrefrn.S) node[ground] {};
        \draw (Qrefri.S) node[currarrow] {} -- ++(-0.5, 0) node[anchor=east] {$I_{in}$};
        \draw (Soma-Igain) node[anchor=east] {$I_{gain}$};
        \draw (Soma-Itau) node[anchor=east] {$I_{tau}$};
        \draw (amp.+) node[currarrow] {} -- ++(-0.5, 0) node[anchor=east] {$I_{\mathrm{spkthr}}$};
        \draw (REFR-Vrefractory) node[anchor=east] {$V_{\mathrm{refractory}}$};
    \end{tikzpicture}}
    \caption{Thresholded integrate and fire circuit.}
    \label{fig:thr_circuit}
\end{figure}

\item Exponential I\&F model

The exponential integrate and fire circuit is shown in Fig.~\ref{fig:exp_circuit}. As the membrane voltage $V_{mem}$ increases and exceeds a certain threshold, a positive feedback current proportional to $I_{mem}$ is injected onto the membrane capacitor.
This makes the neuron fire with an exponential curve as shown in the bottom plots in Fig.~\ref{fig:comparison-of-two-somatic-models-in-different-operation-regimes}.
The threshold is the point at which the exponential feedback overpowers the leak, and is not controlled by any additional parameter.
Measurement results show that the full soma consumes $300pJ$ per somatic spike for 80\,Hz spiking, double the power consumption of the thresholded model (also including the integration power consumption).
The main reason is that the spike pulses are longer.


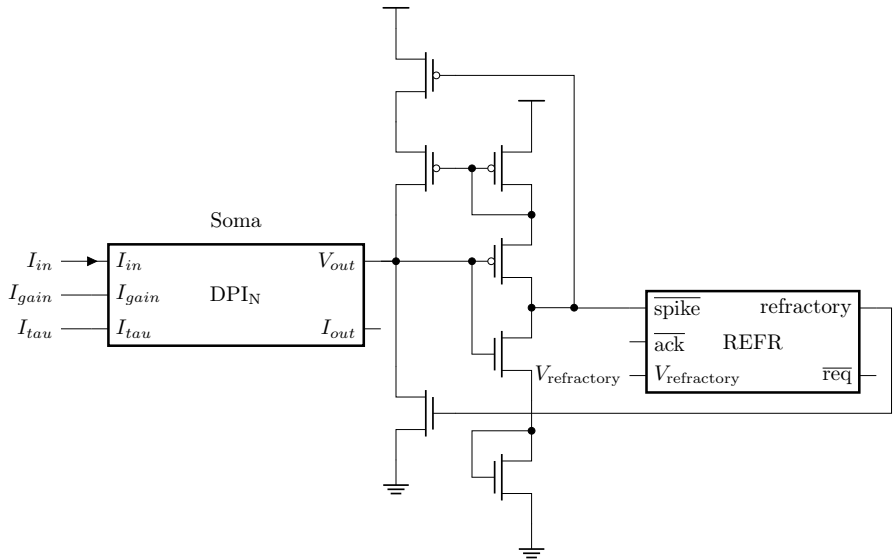
\begin{figure}[!htbp]
    \centering
    \scalebox{.8}{
    \begin{tikzpicture}
        \pic[pic text=DPI\textsubscript{N}] (Soma) at (0, 0) {DPI};
        \draw (0, 1.25) node {Soma};
        \pic [pic text=REFR] (REFR) at(8.5, -0.775) {REFR};
        \draw (Soma-Vout) -| ++(0.25, 0.77) node[pmos, xscale=-1, anchor=D] (Qrefrp) {};
        \draw (Qrefrp.S) node[pmos, xscale=-1, anchor=D] (Qspike) {};
        \draw (Soma-Vout) ++(0.25, 0) to[short, *-] ++(0, -1.75) node[nmos, xscale=-1, anchor=D] (Qrefrn) {};
        \draw (REFR-refractory) -- ++(0.25, 0) |- (Qrefrn.G);
        \draw (Qspike.S) node[rground, yscale=-1] {};
        \draw (Qrefrn.S) node[ground] {};
        \draw (Soma-Vout) -- ++(1.5, 0) node[pmos, anchor=G] (Qfbp1) {};
        \draw (Qfbp1.D) node[nmos, anchor=D] (Qfbn1) {};
        \draw (Qfbn1.S) to[short, -*] ++(0, -0.5) node[nmos, anchor=D] (Qfbn2) {};
        \draw (Qfbp1.S) node[pmos, anchor=D] (Qfbp2) {};
        \draw (Qrefrp.G) to[short, -*] (Qfbp2.G) |- (Qfbp2.D) to[short, *-] (Qfbp1.S);
        \draw (Qfbp1.G) to[short, *-] (Qfbn1.G);
        \draw (Qfbn1.D) to[short, *-*] ++(0.7, 0) |- (REFR-spike);
        \draw (Qfbn1.D) ++(0.7, 0) |- (Qspike.G);
        \draw (Qfbp2.S) node[rground, yscale=-1] {};
        \draw (Qfbn2.S) node[ground] {};
        \draw (Qfbn2.G) |- (Qfbn2.D);
        \draw (Soma-Iin) node[currarrow] {} -- ++(-0.5, 0) node[anchor=east] {$I_{in}$};
        \draw (Soma-Igain) -- ++(-0.5, 0) node[anchor=east] {$I_{gain}$};
        \draw (Soma-Itau) -- ++(-0.5, 0) node[anchor=east] {$I_{tau}$};
        \draw (REFR-Vrefractory) node[anchor=east] {$V_{\mathrm{refractory}}$};
    \end{tikzpicture}}
    \caption{Exponential integrate and fire circuit.}
    \label{fig:exp_circuit}
\end{figure}

\end{enumerate}

\subsubsection{Neuronal dynamics on a longer timescale}

Beside the relatively fast integrate-and-fire dynamics, biological neurons also have dynamics on longer timescales, such as adaptation and homeostasis, which benefit computation.
The common part of these two mechanisms is that they both use the spikes as negative feedback to regulate the excitability of the neuron itself.
Both are implemented with the LPF from Section \ref{sec:low-pass-filter}, sharing a pulse extender with the input being the neuron spikes and $I_{pw}$ configured using the parameter $\mathsf{SOAD\_PWTAU}$.

\begin{enumerate}
    \item Spike-frequency adaptation

The spike-frequency adaptation circuit prevents the neuron from generating a lot of spikes in a very short time.
The adaptation current is the output of the LPF consisting of the shared pulse extender from Section \ref{sec:basic-pulse-extender} and a non-shared DPI.
This current is subtracted from the dendritic current $I_{dendritic}$ of the soma.
The adaptation function is enabled using the latch $\mathsf{SO\_ADAPTATION}$. The individual controllable parameters are the LPF biases: $I_w$ ($\mathsf{SOAD\_W}$), $I_{gain}$ ($\mathsf{SOAD\_GAIN}$) and $I_{tau}$ ($\mathsf{SOAD\_TAU}$). Figure~\ref{fig:adca} shows measurements of the adaptation of the neuron with constant input. Figure~\ref{fig:adsadc} shows the spike-frequency adaptation measurement with alternating DC input. Notice that the parameters were chosen to give long time constants. In real applications, shorter time constants can reduce the effects of device mismatch.

\begin{figure}[!htbp]
    \centering
    \begin{subfigure}[b]{0.4\textwidth}
         \centering
         \includegraphics[width=\textwidth]{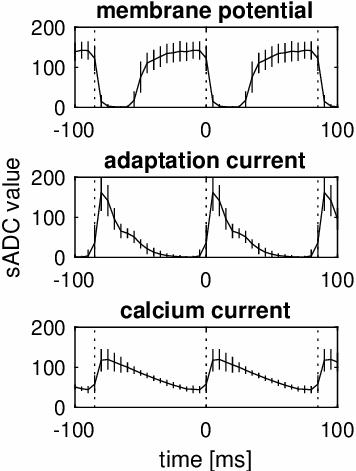}
        \caption{}
         \label{fig:adca}
     \end{subfigure}
    \begin{subfigure}[b]{0.39\textwidth}
         \centering
         \includegraphics[width=\textwidth]{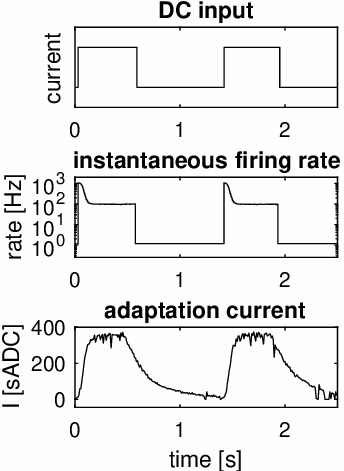}
         \caption{}
         \label{fig:adsadc}
     \end{subfigure}
     \caption{Spike-frequency adaptation and calcium current. (a) Adaptation and calcium currents. The vertical bars are standard deviations over 200 trials. The neuron receives constant DC input. When it fires at time $t=0$, the output of the adaptation and calcium DPIs increase by a certain amount and then decay exponentially. The adaptation current is subtracted from the DC and distal dendritic input . The calcium has a independent weight and a longer time constant, and will fluctuate around a level proportional to the average firing rate of the neuron. (b) Spike-frequency adaptation application example. When DC input is first presented at time $t=0$, the neuron starts to spike at a high rate, causing the adaptation current to increase until it reaches a high enough value to shunt the input, which enters the firing pattern as shown in Fig.~\ref{fig:adsadc}, and the firing rate drops. When the DC input is removed at $t=0.6~\mathrm{s}$, the adaptation current decays exponentially to 0, until the neurons starts firing again (at high rate) when DC input is again presented at $t=1.4~\mathrm{s}$.}
\end{figure}


\item Homeostasis

The homeostasis mechanism is also known as synaptic scaling. It regulates the excitability of the neuron so that the firing rate stays in the medium range (or a target). On <chip-name>, this is achieved with the automatic gain control (AGC) mechanism which can achieve a very long timescale of up to hours.

First, the firing rate of the neuron is estimated using a `calcium current' $I_{Ca}$, which is implemented using an LPF consisting of the pulse extender shared with the spike-frequency adaptation mechanism described above and a non-shared DPI, and should have a relatively long time constant in order to act as an indicator of the overall neural activity.
The calcium current monitored with sADC is shown in Fig.~\ref{fig:adca}), 

The homeostasis function is enabled using the latch $\mathsf{HO\_ENABLE}$. The DPI biases are the weight $I_{Ca, w}$ ($\mathsf{SOCA\_W}$), threshold $I_{Ca, thr}$ ($\mathsf{SOCA\_GAIN}$) and time constant $I_{Ca, tau}$ ($\mathsf{SOCA\_TAU}$). The output (in current mode) is used as an input to the AGC circuit, and can also be chosen as the reversal potential for the conditional conductance dendrites (see Section \ref{sec:conductance-dendrites})

The basic control logic of the AGC is a negative feedback on the somatic gain (or on NMDA gain, controlled by the latch $\mathsf{HO\_SO\_DE}$ where default 0 $=$ somatic, 1 $=$ NMDA) to keep the calcium current around a reference level $I_{Ca, ref}$ ($\mathsf{SOHO\_VREF}$ parameter). Usually,
\begin{equation}
    \frac{\mathrm{d} V_{gain}}{\mathrm{d} t} = \Delta \cdot \mathrm{sign} \left(I_{Ca, ref} - I_{Ca}\right)
\end{equation}
\begin{equation}
    \Delta_+ = \frac{\mathsf{SOHO\_VREF\_H}}{\mathsf{SOHO\_VREF\_M}} = \frac{\mathsf{SOHO\_VREF\_M}}{\mathsf{SOHO\_VREF\_L}} = \Delta_-
\end{equation}
but the ratios could also be set differently to get different ramp-up and ramp-down rates. The output gain voltage can also be reset directly to $\mathsf{SOHO\_VREF\_M}$, which is controlled by the latch $\mathsf{HO\_ACTIVE}$ (default 0 $=$ reset, 1 $=$ enable homeostasis).


Figure \ref{fig:homeostasis-sadc} shows the working mechanism and measurement results of the homeostasis circuit.

\begin{figure}[!htbp]
    \centering
    \begin{subfigure}[b]{0.45\textwidth}
        \centering
        \includegraphics[width=\textwidth]{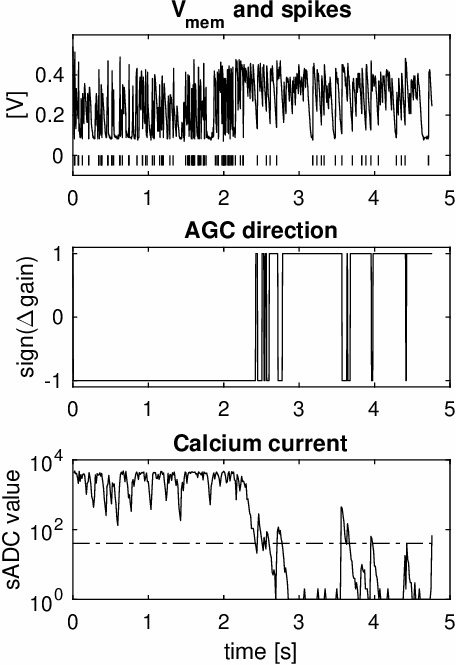}
        \caption{}
    \end{subfigure}
    \begin{subfigure}[b]{0.35\textwidth}
        \centering
        \includegraphics[width=\textwidth]{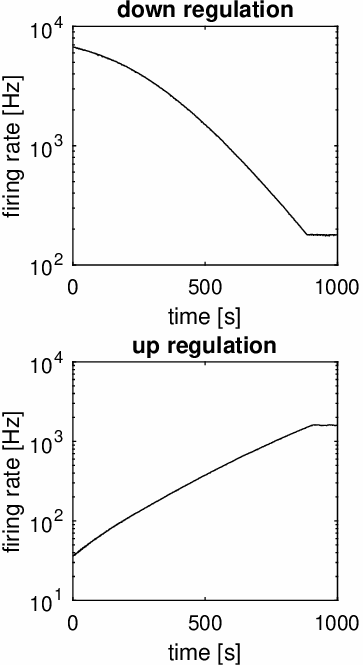}
        \caption{Homeostasis timescale}
    \end{subfigure}  
    \caption{Homeostasis. (a) The neuron receives Poisson-distributed input events at an average of 100\,Hz starting at $t=0$. To begin with, the neuron has a very high gain and thus a very high firing rate. This makes the calcium current $I_{Ca}$ much higher than the reference (target value -- dashed line) and a down regulation of the gain takes place. At around $t=2.5\,\mathrm{s}$, the gain is low enough that the firing rate decreases and the calcium current drops below the reference value, and the gain regulation changes sign. The feedback regulation then keeps the firing activity (calcium current) fluctuating around the reference level. (b) Homeostasis dynamics on a longer timescale. The automatic gain control regulates the gain of the soma very slowly until the firing rate reaches the target in about 15 minutes. Both shorter (milliseconds to seconds) and longer time constants (hours to days) can also be achieved.}
    \label{fig:homeostasis-sadc}
\end{figure}

\end{enumerate}










\subsection{Synaptic compartment} \label{sec:synaptic-compartment}

Each neuron contains 64 synapses and four dendritic branches. 
Each synapse can be attached to any one of the four dendritic compartments.
More details of the dendrite circuits will be discussed in Section \ref{sec:dendritic-compartment}.

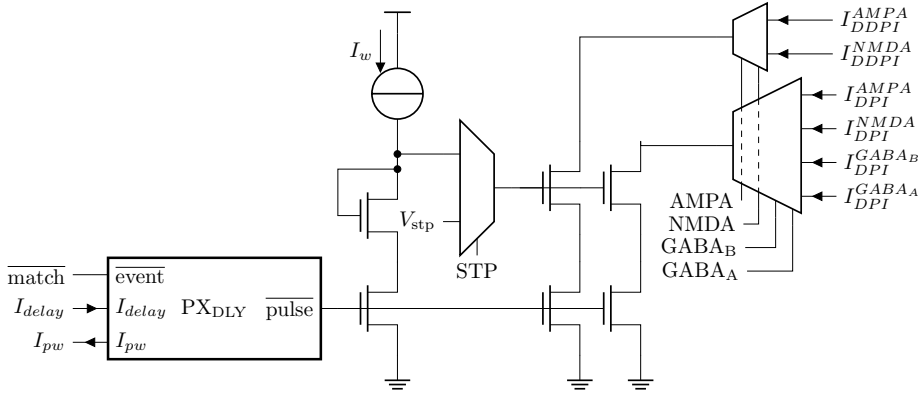
\begin{figure}[!htbp]
    \centering
    \scalebox{0.8}{
    \begin{circuitikz}
        \pic[pic text=PX\textsubscript{DLY}] (PX) at  (0, 0) {PXdly};
        \draw (PX-event) -- ++(-0.3, 0) node[anchor=east] {$\overline{\mathrm{match}}$};
        \draw (PX-Idelay) node[currarrow] {} -- ++(-0.3, 0) node[anchor=east] {$I_{delay}$};
        \draw (PX-Ipw) node[currarrow, xscale=-1] {} -- ++(-0.3, 0) node[anchor=east] {$I_{pw}$};
        \draw (PX-pulse) node[nmos, anchor=G] (Q1) {};
        \draw (Q1.D) node[nmos, anchor=S] (Q4) {};
        \draw (Q1.G) -- ++(3, 0) node[nmos, anchor=G] (Q2) {};
        \draw (Q2.G) -- ++(1, 0) node[nmos, anchor=G] (Q3) {};
        \draw (Q2.D) -- ++(0, 0.46) node[nmos, anchor=S] (Q5) {};
        \draw (Q3.D) -- ++(0, 0.46) node[nmos, anchor=S] (Q6) {};
        \draw (Q1.S) node[ground] {};
        \draw (Q2.S) node[ground] {};
        \draw (Q3.S) node[ground] {};
        \draw (Q4.G) |- (Q4.D) |- ++(0.75, 0.25) node[muxdemux, muxdemux def={Lh=4, Rh=2, NL=2, NB=1, NR=1, w=1}, anchor=lpin 1] (mux) {};
        \draw (mux.rpin 1) -- (Q5.G) -- (Q6.G);
        \draw (mux.lpin 2) node[anchor=east] {$V_{\mathrm{stp}}$};
        \draw (Q4.D) to[short, *-*] ++(0, 0.25) to[current source, f<=$I_w$] ++(0, 2) node[rground, yscale=-1] {};
        \draw (mux.bpin 1) node[anchor=north] {STP};
        \draw (Q6.D) |- (8.25, 2.7) node[muxdemux, muxdemux def={Lh=2, Rh=4, NL=1, NR=4, NB=4, w=2}, anchor=lpin 1] (demux1) {};
        \draw (Q5.D) |- (8.25, 4.5) node[muxdemux, muxdemux def={Lh=1, Rh=2, NL=1, NR=2, NB=2, w=1}, anchor=lpin 1] (demux2) {};
        \draw (demux1.bpin 1) |- ++(0, 0) node[anchor=east] {AMPA};
        \draw (demux1.bpin 2) |- ++(-0.25, -0.25) node[anchor=east] {NMDA};
        \draw (demux1.bpin 3) |- ++(-0.5, -0.5) node[anchor=east] {GABA\textsubscript{B}};
        \draw (demux1.bpin 4) |- ++(-0.75, -0.75) node[anchor=east] {GABA\textsubscript{A}};
        \draw (demux2.bpin 1) -- ++(0, -0.5);
        \draw[dashed] (demux2.bpin 1) ++(0, -0.5) -- (demux1.bbpin 1);
        \draw (demux2.bpin 2) -- ++(0, -0.25);
        \draw[dashed] (demux2.bpin 2) ++(0, -0.25) -- (demux1.bbpin 2);
        \draw (demux1.rpin 1) node[currarrow, xscale=-1] {} -- ++(0.3, 0) node[anchor=west] {$I_{DPI}^{AMPA}$};
        \draw (demux1.rpin 2) node[currarrow, xscale=-1] {} -- ++(0.3, 0) node[anchor=west] {$I_{DPI}^{NMDA}$};
        \draw (demux1.rpin 3) node[currarrow, xscale=-1] {} -- ++(0.3, 0) node[anchor=west] {$I_{DPI}^{GABA_B}$};
        \draw (demux1.rpin 4) node[currarrow, xscale=-1] {} -- ++(0.3, 0) node[anchor=west] {$I_{DPI}^{GABA_A}$};
        \draw (demux2.rpin 1) node[currarrow, xscale=-1] {} -- ++(0.8, 0) node[anchor=west] {$I_{DDPI}^{AMPA}$};
        \draw (demux2.rpin 2) node[currarrow, xscale=-1] {} -- ++(0.8, 0) node[anchor=west] {$I_{DDPI}^{NMDA}$};
    \end{circuitikz}}
    \caption{Synapse block diagram. The input pulse is the active low $\overline{\mathrm{match}}$ signal coming from the content addressable memory (CAM) (see Section \ref{sec:sram-aer-cam-routing-scheme}). The output current of the delayed weighted pulse extender (see Section \ref{sec:delayed-pulse-extender}) will be copied and directed to one of the dendritic branches. The weight can either come from a 4-bit DAC of the type described in Section \ref{sec:flexible-dac} (outputs $I_w$, $n=3$) or from the short term plasticity (STP) output ($V_{\mathrm{stp}}$), chosen by the latch $\mathsf{STP}$ (default $0=$ DAC, $1=$ STP). The delay current parameter comes from another 2-bit DAC of the type described in Section \ref{sec:flexible-dac} (output $I_{delay}$, $n=2$ but with always-on $I_0$). The pulse width control $I_{pw}$ is set by the $\mathsf{SYPD\_EXT}$ parameter. The output demultiplexer uses one-hot encoding, where four latches control whether the current goes to each of the four dendritic branch DPIs. For the two excitatory dendrites AMPA and NMDA, there is also a copy of the current provided to the double DPI (DDPI) responsible for producing alpha-function shaped EPSCs (see Section \ref{sec:double-dpi-alpha-function-epsp}).}
    \label{fig:synapse_block_diagram}
\end{figure}

The synaptic and dendritic compartment generate post-synaptic currents from pre-synaptic events.
The synapse is a delayed, weighted, low-pass-filter as shown in Fig.~\ref{fig:low-pass-filter} that takes the pre-synaptic events as its input and outputs analog pulses with programmable width and height, which are used as the inputs to the dendritic DPI blocks.
A block diagram of the synapse is shown in Fig.~\ref{fig:synapse_block_diagram}.

\subsubsection{Synaptic delay} \label{sec:synaptic-delay}

The delay current DAC of the type described in Section \ref{sec:flexible-dac} contains two digital latches named $\mathsf{precise\_delay}$ for $x_1$ and $\mathsf{mismatched\_delay}$ for $x_2$, and three analog parameters: $\mathsf{SYPD\_DLY0}$ for $I_0$, $\mathsf{SYPD\_DLY1}$ for $I_1$ and $\mathsf{SYPD\_DLY2}$ for $I_2$.
The naming `$\mathsf{precise\_delay}$' and `$\mathsf{mismatched\_delay}$' comes from the design feature that $\mathsf{SYPD\_DLY2}$ has higher mismatch than the other two, in order to give a distribution of delays across a core.
$x_0$ is fixed to 1, which means $\mathsf{SYPD\_DLY0}$ sets the minimum output current and thus maximum delay time.
Different combinations of the settings of the two latches can also be interpreted as providing four groups of delays, as shown in Table \ref{tab:delay-table}.

\begin{table}[!htbp]
    \centering
    \caption{Latch configuration for four groups of delays.}
    \begin{tabular}{|c|c|c|}
        \hline
         & $\mathsf{mismatched\_delay} = 0$ & $\mathsf{mismatched\_delay} = 1$ \\ \hline
        \multirow{2}*{$\mathsf{precise\_delay} = 0$} & $T_{delay} \propto \left(I_{dly0}\right)^{-1}$ & $T_{delay} \propto \left(I_{dly0} + I_{dly2}\right)^{-1}$ \\
         & high delay, low mismatch & high delay, high mismatch \\ \hline
        \multirow{2}*{$\mathsf{precise\_delay} = 1$} & $T_{delay} \propto \left(I_{dly0} + I_{dly1}\right)^{-1}$ & $T_{delay} \propto \left(I_{dly0} + I_{dly1} + I_{dly2}\right)^{-1}$ \\
         & low delay, low mismatch & low delay, high mismatch \\ \hline
    \end{tabular}
    \label{tab:delay-table}
\end{table}



\begin{figure}[!htbp]
    \centering
    \includegraphics[width=0.8\textwidth]{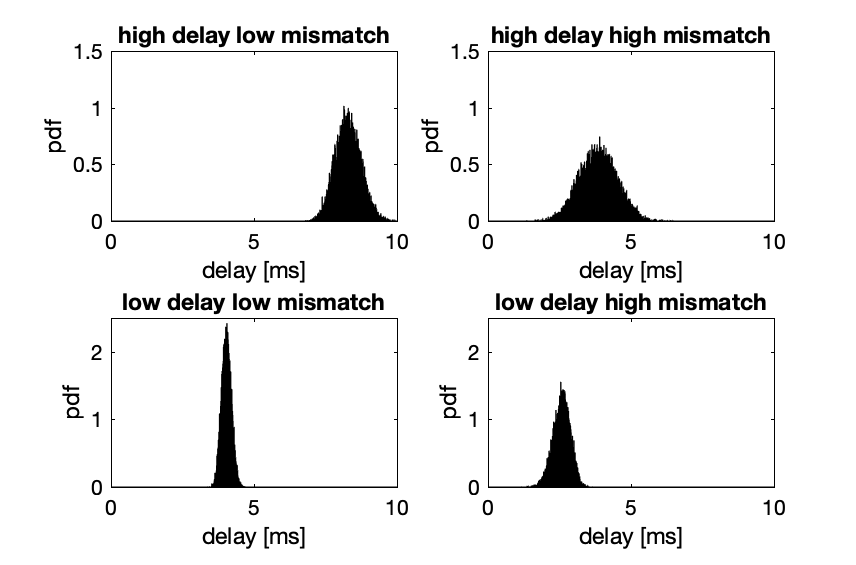}
    \caption{Four groups of synaptic delay distributions. The configuration of the latches are given in Table~\ref{tab:delay-table}. The measurement results show the standard deviations in $I_{dly0}$, $I_{dly1}$ and $I_{dly2}$ to be 5.4\%, 6.7\% and 37.1\% respectively. With the different standard deviations the spread and position of the delay distribution can be freely configured via parameters as $I_{dly0}$ and  $I_{dly1}$ and/or $I_{dly2}$ are summed depending on the individual synaptic configuration. The summed current then controls the effective delay applied.}
    \label{fig:four-synaptic-delay-groups}
\end{figure}

An illustration of the four groups of delay distributions is shown in Fig.~\ref{fig:four-synaptic-delay-groups}.
Note that this is just one example of the analog parameter configurations, shorter (down to a few microseconds) and longer (up to one second) delays are also possible; the combined use of the two precise and one mismatched delay parameters gives control over shaping the delay distribution for the desired application.

\subsubsection{Short-term plasticity} \label{sec:short-term-plasticity}

\begin{figure}[!htbp]
    \centering
    \begin{subfigure}[b]{0.58\textwidth}
         \centering
         \scalebox{0.8}{
            \begin{circuitikz}
                \draw (0, 0) node[nmos] (Qspike) {};
                \draw (Qspike.S) node[ground] {};
                \draw (Qspike.G) node[anchor=east] {pre-spike};
                \draw (Qspike.D) node[nmos, anchor=S] (Qstr) {};
                \draw (Qstr.D) to[short, -*] ++(0, 0) node[pmos, bulk, anchor=D, xscale=-1] (Qr) {};
                \draw (Qr.S) node[anchor=west] {$V_{\mathrm{stpw}}$} |- ++(-0.25, 0) to[short, *-*] ++(-0.25, 0) node[op amp, anchor=out] (buf) {};
                \draw (Qr.S) ++(-0.25, 0) |- (Qr);
                \draw (Qr.S) ++(-0.5, 0) -- ++(0, 1.5) -| (buf.-);
                \draw (buf.+) to[short, -*] ++(-0.5, 0) node[nmos, anchor=G, xscale=-1] (Qin) {};
                \draw (Qin.G) |- (Qin.D) to[current source, *-, f<=$I_{\mathrm{stpw}}$] ++(0, 2) node[rground, yscale=-1] {};
                \draw (Qin.S) node[ground] {};
                \draw (Qr.D) -| (Qr.G) to[short, *-] ++(1, 0) node[anchor=west] {$V_{\mathrm{stp}}$} to[C=$C$] ++(0, -2) node[ground] {};
                \draw (Qstr.G) to[short, -*] ++(-4, 0) node[nmos, xscale=-1, anchor=G] (Qstrin) {};
                \draw (Qstrin.S) node[ground] {};
                \draw (Qstrin.G) |- (Qstrin.D) to[current source, *-, f<=$I_{\mathrm{stpstr}}$] ++(0, 2) node[rground, yscale=-1] {};
            \end{circuitikz}}
         \caption{}
         \label{fig:stdcircuit}
     \end{subfigure}
    \begin{subfigure}[b]{0.4\textwidth}
         \centering
         \includegraphics[width=\textwidth]{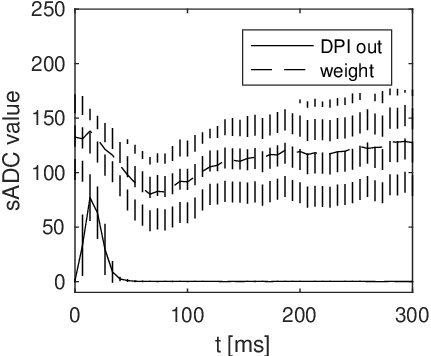}
         \caption{}
         \label{fig:stpsadc}
     \end{subfigure}
     \caption{(a) Short term depression circuit and (b) measurement result. The synapse receives 167\,Hz input from time $t=0$ to $t=60~\mathrm{ms}$, during which the DPI output first increases due to the time constant of several ms, and then decreases due to the decreased weight caused by the short-term depression. After 60\,ms, when there are no further input spikes, the weight recovers with a time constant of roughly 50\,ms. The vertical dashed lines show the standard deviation over 100 trials.}
\end{figure}
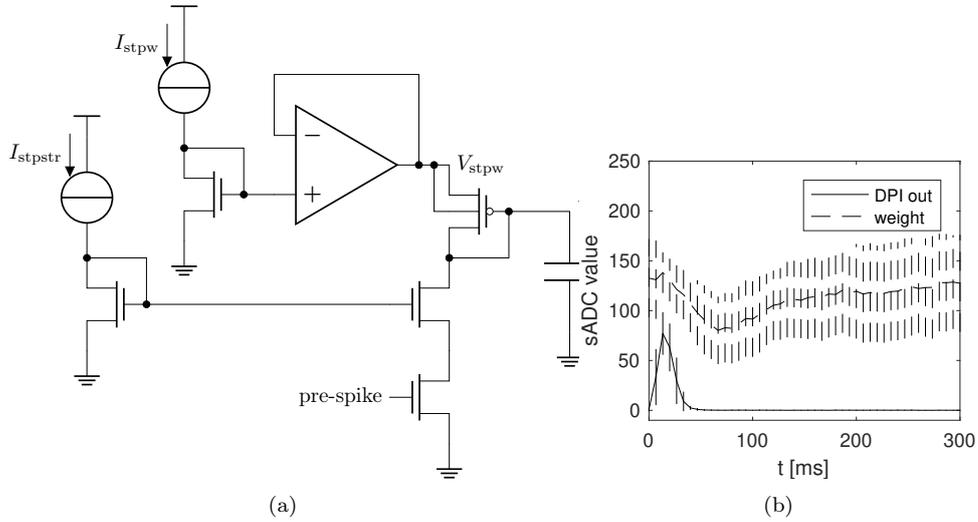

Short-term plasticity (STP) implements depression of the synaptic weight after every pre-synaptic spike.
The circuit is shown in Fig.~\ref{fig:stdcircuit}.
There are two configurable parameters: $I_{\mathrm{stpw}}$ ($\mathsf{SYAN\_STDW}$ parameter) sets the steady state value, and $I_{\mathrm{stpstr}}$ ($\mathsf{SYAN\_STDSTR}$ parameter) sets the strength or how much the output will change for each spike. 

When there is no pre-spike, assuming $V_{\mathrm{stp}}$ is not very far from $V_{\mathrm{stpw}}$, the P-FET is approximately a pseudo-resistor:
\begin{equation}
    C\frac{\mathrm{d}V_{\mathrm{stp}}}{\mathrm{d}t} = \frac{V_{\mathrm{stpw}} - V_{\mathrm{stp}}}{R}
\end{equation}
which means that $V_{\mathrm{stp}}$ will converge to $V_{\mathrm{stpw}}$ exponentially with time constant $\tau=RC$.
For small signals ($V_{\mathrm{stp}} \approx V_{\mathrm{stpw}}$) the corresponding current $I_{\mathrm{stp}}$ following $I_{\mathrm{stp}} = I_0 e^{\kappa\frac{V_{\mathrm{stp}}}{U_T}}$ also converges at $\tau$.

During a pre-spike pulse, assuming $I_{\mathrm{stpstr}} \gg I_0 e^{\kappa\frac{V_{\mathrm{stpw}} - V_{\mathrm{stp}}}{U_T}}$:
\begin{equation}
    C\frac{\mathrm{d}V_{\mathrm{stp}}}{\mathrm{d}t} = - I_{\mathrm{stpstr}}
\end{equation}
which means that $V_{\mathrm{stp}}$ will drop linearly at rate $\frac{I_{\mathrm{stpstr}}}{C}$, and the output current $I_{\mathrm{stp}}$ decays exponentially with time constant $\tau=\frac{C U_T}{\kappa I_{\mathrm{stpstr}}}$.

\subsection{Dendritic compartment} \label{sec:dendritic-compartment}

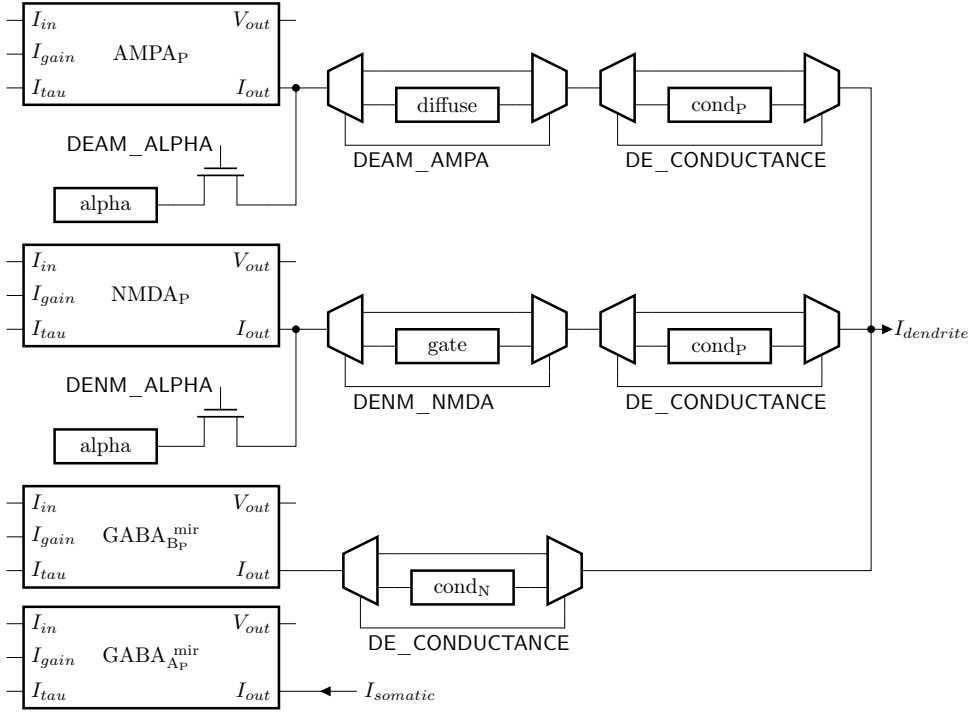
\begin{figure}[!htbp]
    \centering
    \scalebox{0.8}{
    \begin{circuitikz}
        \ctikzset{multipoles/thickness=3}
        \pic[pic text=AMPA\textsubscript{P}] (AMPAe) at (0, 10) {DPI};
        \draw (-0.75, 7.5) node[dipchip, num pins=2, draw only pins={2}, hide numbers, no topmark] (AMPAalpha) {alpha};
        \pic[pic text=NMDA\textsubscript{P}] (NMDAe) at (0, 6) {DPI};
        \draw (-0.75, 3.5) node[dipchip, num pins=2, draw only pins={2}, hide numbers, no topmark] (NMDAalpha) {alpha};
        \pic[pic text=$\mathrm{GABA_{B_{P}}^{\ mir}}$] (GABAB) at (0, 2) {DPI};
        \pic[pic text=$\mathrm{GABA_{A_{P}}^{\ mir}}$] (GABAA) at (0, 0) {DPI};
        
        \draw (AMPAalpha.pin 2) node[nmos, anchor=S, rotate=-90] (Qampa) {};
        \draw (Qampa.G) node[anchor=east] {$\mathsf{DEAM\_ALPHA}$};
        \draw (Qampa.D) -| (AMPAe-Iout);
        \draw (AMPAe-Iout) to[short, *-] ++(0.25, 0) node[muxdemux, muxdemux def={Lh=1, Rh=2, NL=1, NB=1, NR=2, w=1}, anchor=lpin 1] (AMPAmux1) {};
        \draw (AMPAmux1.rpin 2) node[dipchip, num pins=2, hide numbers, no topmark, anchor=pin 1] (AMPAdiff) {diffuse};
        \draw (AMPAdiff.pin 2) node[muxdemux, muxdemux def={Lh=2, Rh=1, NL=2, NB=1, NR=1, w=1}, anchor=lpin 2] (AMPAdemux1) {};
        \draw (AMPAmux1.rpin 1) -- (AMPAdemux1.lpin 1);
        \draw (AMPAmux1.bpin 1) -- ++(0, -0.25) node[anchor=north west] {$\mathsf{DEAM\_AMPA}$} -| (AMPAdemux1.bpin 1);
        \draw (AMPAdemux1.rpin 1) node[muxdemux, muxdemux def={Lh=1, Rh=2, NL=1, NB=1, NR=2, w=1}, anchor=lpin 1] (AMPAmux2) {};
        \draw (AMPAmux2.rpin 2) node[dipchip, num pins=2, hide numbers, no topmark, anchor=pin 1](AMPAcond){cond\textsubscript{P}};
        \draw (AMPAcond.pin 2) node[muxdemux, muxdemux def={Lh=2, Rh=1, NL=2, NB=1, NR=1, w=1}, anchor=lpin 2] (AMPAdemux2) {};
        \draw (AMPAmux2.rpin 1) -- (AMPAdemux2.lpin 1);
        \draw (AMPAmux2.bpin 1) -- ++(0, -0.25) node[anchor=north west] {$\mathsf{DE\_CONDUCTANCE}$} -| (AMPAdemux2.bpin 1);
        
        \draw (NMDAalpha.pin 2) node[nmos, anchor=S, rotate=-90] (Qnmda) {};
        \draw (Qnmda.G) node[anchor=east] {$\mathsf{DENM\_ALPHA}$};
        \draw (Qnmda.D) -| (NMDAe-Iout);
        \draw (NMDAe-Iout) to[short, *-] ++(0.25, 0) node[muxdemux, muxdemux def={Lh=1, Rh=2, NL=1, NB=1, NR=2, w=1}, anchor=lpin 1] (NMDAmux1) {};
        \draw (NMDAmux1.rpin 2) node[dipchip, num pins=2, hide numbers, no topmark, anchor=pin 1] (NMDAdiff) {gate};
        \draw (NMDAdiff.pin 2) node[muxdemux, muxdemux def={Lh=2, Rh=1, NL=2, NB=1, NR=1, w=1}, anchor=lpin 2] (NMDAdemux1) {};
        \draw (NMDAmux1.rpin 1) -- (NMDAdemux1.lpin 1);
        \draw (NMDAmux1.bpin 1) -- ++(0, -0.25) node[anchor=north west] {$\mathsf{DENM\_NMDA}$} -| (NMDAdemux1.bpin 1);
        \draw (NMDAdemux1.rpin 1) node[muxdemux, muxdemux def={Lh=1, Rh=2, NL=1, NB=1, NR=2, w=1}, anchor=lpin 1] (NMDAmux2) {};
        \draw (NMDAmux2.rpin 2) node[dipchip, num pins=2, hide numbers, no topmark, anchor=pin 1](NMDAcond){cond\textsubscript{P}};
        \draw (NMDAcond.pin 2) node[muxdemux, muxdemux def={Lh=2, Rh=1, NL=2, NB=1, NR=1, w=1}, anchor=lpin 2] (NMDAdemux2) {};
        \draw (NMDAmux2.rpin 1) -- (NMDAdemux2.lpin 1);
        \draw (NMDAmux2.bpin 1) -- ++(0, -0.25) node[anchor=north west] {$\mathsf{DE\_CONDUCTANCE}$} -| (NMDAdemux2.bpin 1);
        
        \draw (GABAB-Iout) -- ++(0.5, 0) node[muxdemux, muxdemux def={Lh=1, Rh=2, NL=1, NB=1, NR=2, w=1}, anchor=lpin 1] (GABAmux) {};
        \draw (GABAmux.rpin 2) node[dipchip, num pins=2, hide numbers, no topmark, anchor=pin 1](GABAcond){cond\textsubscript{N}};
        \draw (GABAcond.pin 2) node[muxdemux, muxdemux def={Lh=2, Rh=1, NL=2, NB=1, NR=1, w=1}, anchor=lpin 2] (GABAdemux) {};
        \draw (GABAmux.rpin 1) -- (GABAdemux.lpin 1);
        \draw (GABAmux.bpin 1) -- ++(0, -0.25) node[anchor=north west] {$\mathsf{DE\_CONDUCTANCE}$} -| (GABAdemux.bpin 1);
        
        \draw (GABAA-Iout) -- ++(0.5, 0) node[currarrow, xscale=-1] {} -- ++(0.5, 0) node[anchor=west] {$I_{somatic}$};

        \draw (AMPAdemux2.rpin 1) -- ++(0.25, 0) |- (GABAdemux.rpin 1);
        \draw (NMDAdemux2.rpin 1) to[short, -*] ++(0.25, 0) -- ++(0.25, 0) node[currarrow] {} node[anchor=west] {$I_{dendrite}$};

    \end{circuitikz}}
    \caption{Dendrite compartment block diagram. Input $I_{in}$ is the sum of the delayed extended weighted pulse coming from the synaptic compartments. The two excitatory dendrites AMPA and NMDA have conditional alpha function blocks, and along with GABA\textsubscript{B} have conditional conductance blocks. The AMPA dendrite also contains a conditional diffusion block. The NMDA dendrite has conditional membrane voltage gating block. The conditional blocks can be bypassed using digital latches. The sum of the output currents from AMPA($+$), NMDA($+$) and GABA\textsubscript{B}($-$) is $I_{dendritic}$; the only component of $I_{somatic}$ is the shunting current into GABA\textsubscript{A}.}
    \label{fig:dendrite-block-diagram}
\end{figure}

The dendritic block contains two excitatory (AMPA and NMDA) and two inhibitory (GABA\textsubscript{B} and GABA\textsubscript{A}) DPI compartments, which turn pre-synaptic events into excitatory and inhibitory PSCs. The block diagram is shown in Fig.~\ref{fig:dendrite-block-diagram}.

\subsubsection{Conductance dendrites} \label{sec:conductance-dendrites}

\begin{figure}[!htbp]
    \centering
    \scalebox{0.8}{
    \begin{circuitikz}
        \draw (0, 0) node[pmos] (Qrev) {};
        \draw (Qrev.D) to[short, -*] ++(0, -0.5) node[nmos, xscale=-1, anchor=D] (Qn1) {};
        \draw (Qn1.S) node[ground] {};
        \draw (Qn1.D) -| (Qn1.G) to[short, *-] ++(0.25, 0) node[nmos, anchor=G] (Qn2) {};
        \draw (Qn2.S) node[ground] {};
        \draw (Qn2.D) to[short, -*] ++(0, 0.25) -- ++(0, 0.25) node[pmos, xscale=-1, anchor=D] (Qv) {};
        \draw (Qrev.S) -- (Qv.S);
        \draw ($(Qrev.S)!0.5!(Qv.S)$) to[current source, *-, f<=$I_{in}$] ++(0, 2) node[rground, yscale=-1] {};
        \draw (Qn2.D) ++(0, 0.25) -- ++(0.5, 0) node[currarrow] {} node[anchor=west] {$I_{out}$};
        \draw (Qrev.G) node[anchor=east] {$V_{reversal}$};
        \draw (Qv.G) node[anchor=west] {$V_{neuron}$};
        \ctikzset{multipoles/thickness=3}
        \draw ($(Qn1.S)!0.5!(Qn2.S)$) ++(0, -1.5) node[dipchip, num pins=2, hide numbers, no topmark] (cond) {cond\textsubscript{P}};
        \draw (cond.pin 1) node[currarrow] {} -- ++(-0.5, 0) node[anchor=east] {$I_{in}$};
        \draw (cond.pin 2) node[currarrow] {} -- ++(0.5, 0) node[anchor=west] {$I_{out}$};
    \end{circuitikz}
    \begin{circuitikz}
        \draw (0, 0) node[nmos] (Qrev) {};
        \draw (Qrev.D) to[short, -*] ++(0, 0.5) node[pmos, xscale=-1, anchor=D] (Qp1) {};
        \draw (Qp1.S) node[rground, yscale=-1] {};
        \draw (Qp1.D) -| (Qp1.G) to[short, *-] ++(0.25, 0) node[pmos, anchor=G] (Qp2) {};
        \draw (Qp2.S) node[rground, yscale=-1] {};
        \draw (Qp2.D) to[short, -*] ++(0, -0.25) -- ++(0, -0.25) node[nmos, xscale=-1, anchor=D] (Qv) {};
        \draw (Qrev.S) -- (Qv.S);
        \draw ($(Qrev.S)!0.5!(Qv.S)$) to[current source, *-, f>=$I_{in}$] ++(0, -2) node[ground] (gnd) {};
        \draw (Qp2.D) ++(0, -0.25) -- ++(0.5, 0) node[currarrow, xscale=-1] {} -- ++(0.5, 0) node[anchor=west] {$I_{out}$};
        \draw (Qrev.G) node[anchor=east] {$V_{reversal}$};
        \draw (Qv.G) node[anchor=west] {$V_{neuron}$};
        \draw (gnd) ++(0, -1.5) node[dipchip, num pins=2, hide numbers, no topmark] (cond) {cond\textsubscript{N}};
        \draw (cond.pin 1) node[currarrow, xscale=-1] {} -- ++(-0.5, 0) node[anchor=east] {$I_{in}$};
        \draw (cond.pin 2) node[currarrow, xscale=-1] {} -- ++(0.5, 0) node[anchor=west] {$I_{out}$};
    \end{circuitikz}}
    \caption{Conductance block circuit (left: P-type, right: N-type)}
    \label{fig:conductance-circuit}
\end{figure}
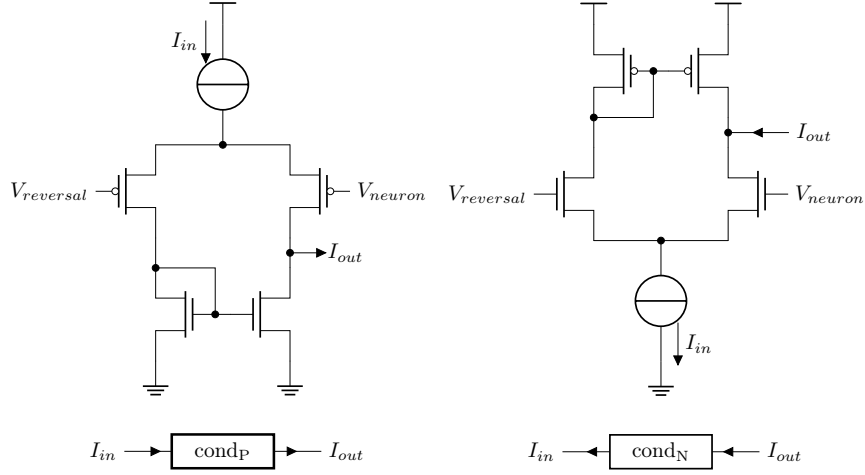

\begin{figure}[!htbp]
    \centering
    \begin{subfigure}[b]{0.32\textwidth}
         \centering
         \includegraphics[width=\textwidth]{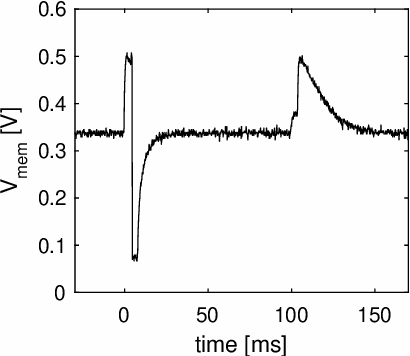}
         \caption{Conductance mode effects}
         \label{fig:conductance-example}
     \end{subfigure}
    \begin{subfigure}[b]{0.32\textwidth}
         \centering
         \includegraphics[width=\textwidth]{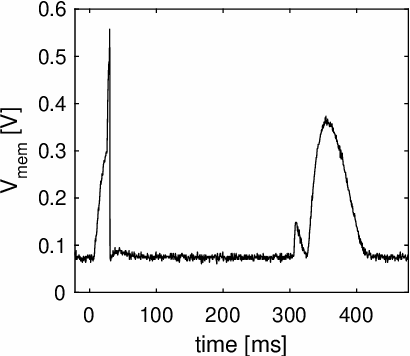}
         \caption{Alpha function effects}
         \label{fig:alpha-example}
     \end{subfigure}
     \begin{subfigure}[b]{0.32\textwidth}
         \centering
         \includegraphics[width=\textwidth]{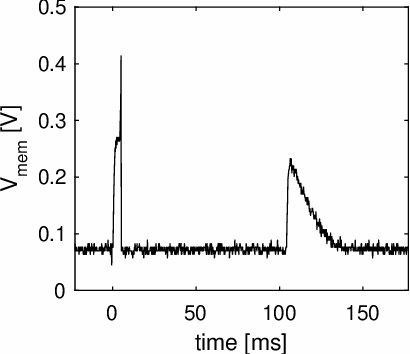}
         \caption{NMDA gating effects}
         \label{fig:nmda-example}
     \end{subfigure}
     \caption{Application examples for the conditional dendritic functions: conductance, alpha-function and NMDA gating. (a) The neuron has both excitatory dendrites in conductance mode, with one of the reversal potentials set to 0.5\,V and its synaptic weight very high, and the other has the reversal potential at around 0.7\,V but the weight is low. The spiking threshold is set to around 0.6\,V. Starting from the resting potential at around 0.35\,V, when the first dendrite receives an input spike at time $t=0$, it charges the soma up to the reversal potential, and when the second dendrite receives a spike shortly afterwards at time $t=5~\mathrm{ms}$, it further charges the soma until it crosses the firing threshold and emits a spike (the neuron then goes into its refractory period). However, if the second dendrite receives its input (at time $t=100~\mathrm{ms}$) before the first dendrite (at time $t=105~\mathrm{ms}$), then since the second dendrite by itself cannot drive the neuron to fire (due to the low weight) but the first dendrite cannot charge the soma once $V_{mem}$ reaches its reversal potential (0.5\,V), which is lower than the firing threshold, the neuron does not emit a spike and slowly leaks back to its resting potential. Thus this neuron could be used to detect the order in time of the two inputs, since it fires if and only if one input comes shortly before the other. (b) The neuron uses both excitatory dendrites, one using the alpha function and the other using only the normal DPI. If the first dendrite receives an input spike at time $t=0$, it will start to charge the soma slowly (according to the alpha function), and if the second dendrite receives a spike shortly afterwards at time $t=20~\mathrm{ms}$, it will charge the soma even further to cross the firing threshold and emit a spike. However, if the second dendrite receives its input (at time $t=300~\mathrm{ms}$) before the first dendrite (at time $t=320~\mathrm{ms}$), then since the effect of the second dendrite goes away very fast, and the first dendrite by itself cannot charge the soma to cross the firing threshold either, the neuron does not emit a spike. This mechanism introduces a delayed dynamic, so it can also be used to detect the order of the two inputs. (c)  The neuron uses the AMPA and NMDA dendrites. If the AMPA dendrite receives an input spike at time $t=0$, it will charge the membrane potential to a value higher than the NMDA threshold (which is set to around 0.1\,V), and if the NMDA dendrite receives a spike shortly afterwards at time $t=5~\mathrm{ms}$, it will charge the soma to cross the firing threshold and emit a spike. However, if the NMDA dendrite receives the input (at time $t=100~\mathrm{ms}$) before the AMPA dendrite (at time $t=105~\mathrm{ms}$), then since the membrane potential at the moment when the NDMA dendrite receives the spike was still lower than the NMDA threshold, it has no effect on the soma, and the AMPA dendrite by itself cannot charge the soma to cross the firing threshold, so the neuron does not emit a spike. This mechanism forces an asymmetric condition on when the soma receives the input, so it can also be used to detect the order of the two inputs.}
\end{figure}

The AMPA, NDMA and GABA\textsubscript{B} dendrites can be individually switched to conductance mode to emulate a large class of biologically inspired synaptic models. The circuit is shown in Fig.~\ref{fig:conductance-circuit} and is adapted from \cite{Sumislawska_etal16}. The output from the conductance block $I_{conductance}$ to the soma is a $\tanh$ function of the difference between the reversal potential $V_{reversal}$ set by the parameter $\mathsf{REV}$ and $V_{neuron}$ which is either the somatic membrane potential $V_{mem}$ or the calcium current $V_{Ca}$ (selected using the latch $\mathsf{COHO\_CA\_MEM}$, default $0 = V_{mem}$, $1=V_{Ca}$):
\begin{equation*}
    I_{conductance} = I_{dendrite} \tanh \frac{V_{reversal} - V_{neuron}}{U_T}
\end{equation*}
The measurement result shown in Fig.~\ref{fig:conductance-example} illustrates a simple example of using the conductance function.

\subsubsection{Double-DPI --- alpha function EPSC} \label{sec:double-dpi-alpha-function-epsp}

Both AMPA and NMDA EPSCs can accurately emulate alpha function synapse potentials with an additional inhibitory DPI (P-type but with mirrored output as described in Section \ref{se:mirrored-output}) \cite{Sumislawska_etal16}. The difference current between the excitatory and the inhibitory DPIs (EDPI and IDPI) produces an alpha-function-shaped EPSP. Quantitatively,
\begin{equation*}
    I_{\mathrm{DDPI}} = \max \left(I_{\mathrm{EDPI}} - I_{\mathrm{IDPI}}, 0 \right) = \max \left( W_{E} e^{-\frac{t}{\tau_E}} - W_{I} e^{-\frac{t}{\tau_I}}, 0 \right)
\end{equation*}
where the coefficients $W_{E}$ and $W_{I}$ are controlled by the parameters $\mathsf{EGAIN}$ and $\mathsf{IGAIN}$ respectively and the time constants $\tau_E$ and $\tau_I$ are controlled by the parameters $\mathsf{ETAU}$ and $\mathsf{ITAU}$ respectively as described in Section \ref{sec:low-pass-filter}. The measurement result shown in Fig.~\ref{fig:alpha-example} illustrates a simple example of using the alpha function dendrite. 




\subsubsection{Diffusion over a 2D grid}

\begin{figure}[!htbp]
    \centering
    \begin{subfigure}[b]{0.6\textwidth}
         \centering
         \scalebox{.8}{
            \begin{circuitikz}
                \foreach \i in {0, 1, 2}
                {
                    \foreach \j in {0, 1, 2}
                    {
                        \begin{scope}[shift={(\i*3+\j*1.8,\j*3)}]
                            \draw (0, 0) to[R=$R_h$] (3, 0);
                            \draw (0, 0) to[R=$R_v$] (1.8, 3);
                            \draw (0, 1) node[anchor=south] {$I_{in}$} -- (0, 0.5) node[currarrow, rotate=-90] {}  -- (0, 0) to[R=$R_n$, *-] ++(0, -1.8) node[currarrow, rotate=-90] {} node[anchor=west] {$I_{out}$} -- ++ (0, -0.2);
                            \draw[dashed] (-0.3, 2.5) node[anchor=north west] {(\i,\j)} -| (1.4, 0.8) -| (2.6, -0.4) -| (0.8, -2.3) -| cycle;
                        \end{scope}
                    }
                }
            \end{circuitikz}}
         \caption{}
        \label{fig:diffusion-circuit}
     \end{subfigure}
     \begin{subfigure}[b]{0.38\textwidth}
          \centering
              \begin{circuitikz}
            \draw (0, 0) node[pmos] (Qp1) {};
            \draw (Qp1.S) to[short, -*] ++(0.5, 0) {} to[current source, f<=$I_{in}$] ++(0, 2) node[rground, yscale=-1] {};
            \draw (Qp1.S) ++(0.5, 0) -- ++(0.5, 0) node[pmos, xscale=-1, anchor=S] (Qp2) {};
            \draw (Qp2.D) node[ground] {};
            \draw (Qp2.G) node[anchor=west] {$V_{m}$};
            \draw (Qp1.G) node[anchor=east] {$V_{\mathrm{NMDA}}$};
            \draw (Qp1.D) node[currarrow, rotate=-90] {} node[anchor=east] {$I_{out}$} -- ++(0, -0.5);
        \end{circuitikz}
            \caption{}
    \label{fig:nmda-circuit}
     \end{subfigure}
    \caption{Conditional dendritic blocks. (a) 2D diffusive grid connected to the AMPA dendrite. This can be enabled neuron-wise using the latch $\mathsf{DEAM\_AMPA}$, and includes the corresponding neuron pseudo-resistor $\mathsf{NRES}$ ($R_n$ in the figure), the horizontal pseudo-resistor $\mathsf{HRES}$ ($R_h$ in the figure, between neuron $n$ and $n+1$), and the vertical pseudo-resistor $\mathsf{VRES}$ ($R_v$ in the figure, between neuron $n$ and $n+16$). The pseudo-resistors are implemented with single P-FETs, and the controllable parameters are the gate voltages $\mathsf{DEAM\_NRES}$, $\mathsf{DEAM\_HRES}$ and $\mathsf{DEAM\_VRES}$. (b) NMDA gating. When enabled using the $\mathsf{DENM\_NMDA}$ latch, the output current of the NMDA DDPI (here $I_{in}$) will flow out into the neuron's $I_{dendritic}$ if and only if the membrane potential $V_{m}$ is higher than the NMDA threshold $V_{\mathrm{NMDA}}$ (set by $\mathsf{DENM\_NMREV}$ parameter).}

\end{figure}
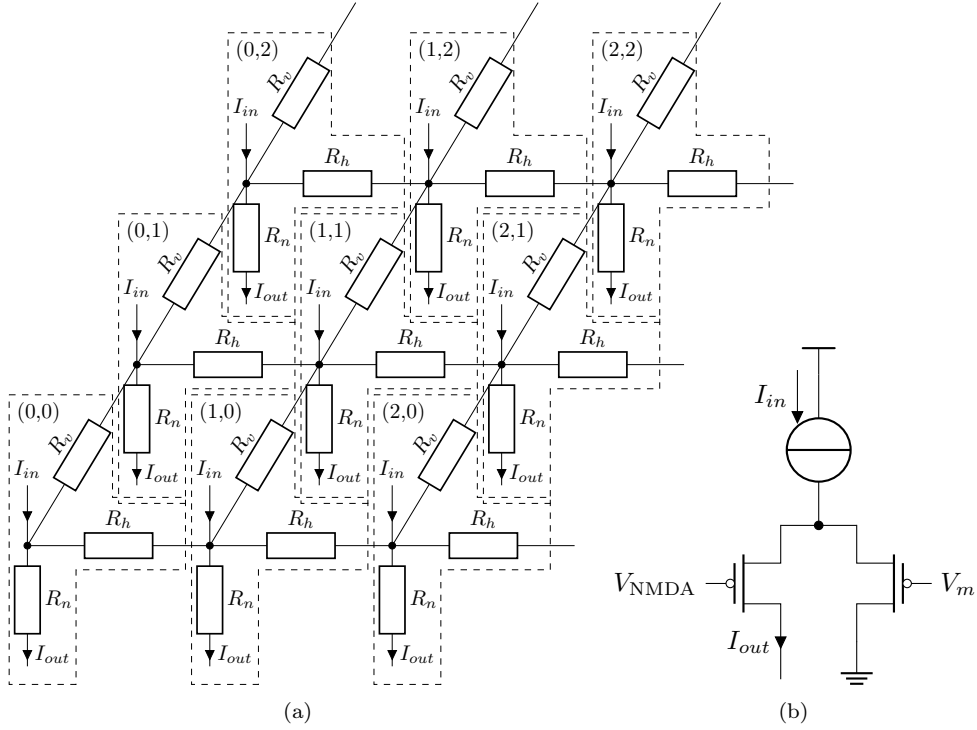

\begin{figure}[!htbp]
    \centering
    \includegraphics[width=\textwidth]{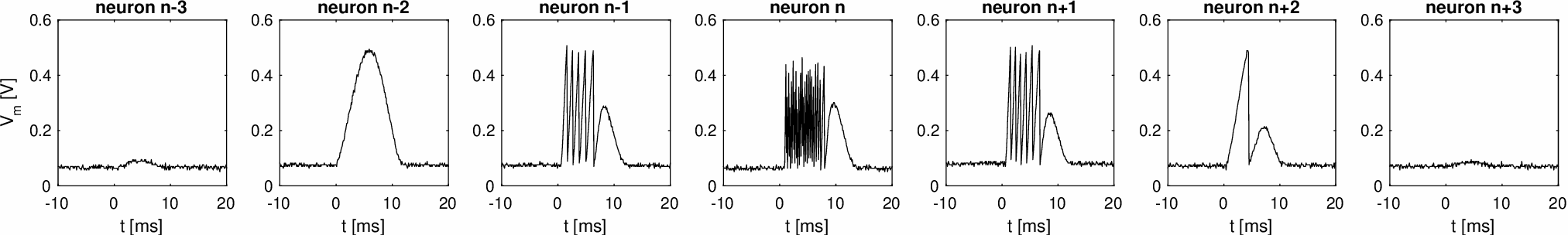}
    \caption{AMPA diffusion in one dimension. The input spike is only sent to the neuron in the middle (neuron n), but the diffusion creates a bump in the membrane potentials in the neurons in its (here, 1D) neighborhood.}
    \label{fig:ampa-diffusion-in-one-dimension}
\end{figure}

The AMPA dendritic compartment offers an conditional 1D or 2D resistive grid similar to that described in \cite{Benjamin_etal14} to diffuse incoming EPSCs between nearby neurons. The circuit is shown in Fig.~\ref{fig:diffusion-circuit}. An example of one dimensional (horizontal) diffusion is shown in Fig.~\ref{fig:ampa-diffusion-in-one-dimension}.

\subsubsection{NMDA --- gating with the membrane potential} \label{sec:nmda-gating-with-membrane-potential}

The NMDA dendritic compartment can gate the incoming current depending on the membrane potential, shown in Fig.~\ref{fig:nmda-circuit}. The measurement result shown in Fig.~\ref{fig:nmda-example} illustrates a simple example of using the NMDA threshold circuit.




It is important to note that disabling the gating using the latch and enabling it but setting $V_{\mathrm{NMDA}}$ to 0 are not equivalent, as one would predict from an ideal computational model, because of the different leakage current with and without the NMDA gating circuit. Measurement shows the latter condition may give several picoamps more leakage thus decreasing the excitability of the neuron.









\section{Digital event routing and mapping scheme} \label{sec:digital-spike-routing-scheme}

\subsection{Inter-neuron routing and connection mapping scheme} \label{sec:sram-aer-cam-routing-scheme}

The routing scheme used within the core has been inspired by \cite{Moradi_etal18}.
The details of this should not concern the user unless special edge cases are encountered (e.g., applications requiring very low latency or very high firing rate or many neurons firing simultaneously).
The user must however understand the addressing scheme in order to make connections between neurons.


The principle idea is to use AER to encode the spikes into a stream of bit patterns, so that they can be easily transmitted and routed within and outside of the chips.
More specifically, on \chipname{}, each normal inter-neuron event is encoded as a 24-bit word comprising a format indicator bit (bit 23 = 0) and four variable fields as shown in Table \ref{tab:aer-words}, the event \texttt{tag}, the target chip displacements in the x and y directions (\texttt{dx} and \texttt{dy} respectively) and the \texttt{cores} mask that determines which cores the event is delivered to on the target chip.

\begin{table}[!htbp]
    \caption{Format of \chipname{} AER event words.}
    \centering
    \tiny
    \resizebox{\textwidth}{!}{
    \begin{tabular}{|l|c|c|c|c|c|c|c|c|c|c|c|c|c|c|c|c|c|c|c|c|c|c|c|c|}
        \hline
        \multicolumn{1}{|r|}{Bit no.} & 23 & 22 & 21 & 20 & 19 & 18 & 17 & 16 & 15 & 14 & 13 & 12 & 11 & 10 & 9 & 8 & 7 & 6 & 5 & 4 & 3 & 2 & 1 & 0 \\
        \hline 
        Inter-neuron event & 0 & \multicolumn{11}{c|}{\texttt{tag}} & \multicolumn{4}{c|}{\texttt{dy}} & \multicolumn{4}{c|}{\texttt{dx}} & \multicolumn{4}{c|}{\texttt{cores}} \\
        \hline
        Sensor event & 1 & pol & \multicolumn{9}{c|}{\texttt{pixel\_y}} & \multicolumn{9}{c|}{\texttt{pixel\_x}} & \multicolumn{2}{c|}{\texttt{dy}} & \multicolumn{2}{c|}{\texttt{dx}} \\
        \hline
    \end{tabular}
    }
    \label{tab:aer-words}
\end{table}

Each neuron has four 23-bit SRAMs to store four combinations of \texttt{tag}, \texttt{dy}, \texttt{dx} and \texttt{cores}.
When the pre-neuron fires, the router will read and transmit the content of these four SRAMs. This is known as source mapping.
This is in contrast to DYNAP-SE \cite{Moradi_etal18} which does not include arbitrary source mapping and is therefore limited in the network connectivity it could implement.
There is no dedicated `enable' bit for an outgoing event, but if none of the four cores on the target chip is selected ($\mathtt{cores} = 0000\mathtt{b}$), the event will be dropped by the router.

The events are transmitted inside a 2D grid of trees, where every chip has a tree routing structure and four grid connections to the neighboring chips.

When an event arrives at a chip (this can be the sender chip itself), the top-level router will decide, based on the target chip displacement bits, whether the event should be kept for this chip ($\mathtt{dx} = 0$ and $\mathtt{dy} = 0$) or forwarded further on one of the grid buses (west if $\mathtt{dx} < 0$, east if $\mathtt{dx} > 0$, south if $\mathtt{dx} = 0$ and $\mathtt{dy} < 0$, north if $\mathtt{dx} = 0$ and $\mathtt{dy} > 0$, see Section \ref{sec:inter-chip-event-communication} for more details).
If the top-level router decides to keep the event, it will be sent to all cores that are selected in the \texttt{cores} bits.


Once it has arrived in a core, an event is identified only by its 11-bit \texttt{tag}.
This means that when two events with the same tag arrive in the same core, there is no way for a neuron in that core to tell them apart, even if they come from different pre-neurons.
This is used to share synapses as the tags can be assigned arbitrarily in the source mapping.
The 11-bit tag is broadcast to all 256 neurons $\times$ 64 synapses in the core.
Each synapse is provided with an 11-bit CAM.
If all eleven bits of the broadcast tag match those in a synapse's CAM, an active low `match' signal is sent to the synapse circuitry as described in the caption of Fig.~\ref{fig:synapse_block_diagram}.
This matching process is known as destination mapping.

\subsection{Example configurations} \label{sec:example-configuration}

To better illustrate the tag scheme, two concrete examples of how the \texttt{tag} in the SRAMs of the pre-neurons and in the CAMs of the post-neurons can be used are shown in Python-like pseudo-code:

\begin{enumerate}
    \item For all-to-all connections from \texttt{n} neurons (in list \texttt{pre}) to \texttt{r} synapses on each of \texttt{n} neurons (in list \texttt{post}), a single tag \texttt{x} is used:
    \begin{verbatim}
    for i in range(n):
        neurons[pre[i]].srams[0].tag = x
        for k in range(r):
            neurons[post[i]].cams[k].tag = x
    \end{verbatim}
    \item To connect each of the \texttt{n} pre neurons (in list \texttt{pre}) to the $(2\mathtt{r} + 1)$ neighbors $\!\!\pmod{\mathtt{n}}$ in the \texttt{n} post neurons (in list \texttt{post}), tags in the interval $[\mathtt{x}, \mathtt{x} + \mathtt{n}]$ are used:
    \begin{verbatim}
    for i in range(n):
        neurons[pre[i]].srams[0].tag = x + i
        for k in range(-r, r + 1):
            neurons[post[i]].cams[r + k].tag = x + ((i + k) % n)
    \end{verbatim}
\end{enumerate}

\subsection{Multiplexing of four neurons}

For networks that require higher synaptic counts, there is an option to merge the dendrites of four neurons into one (enabled using the latch $\mathsf{DE\_MUX}$, set for each core individually). This increases the number of synapses per neuron to 256 and reduces the number of neurons by a factor of four. More specifically, the PSCs $I_{dendritic}$ and $I_{somatic}$ of neurons 0, 1, 16 and 17 will all go to the soma of neuron 0; those of neurons 2, 3, 18 and 19 will go to the soma of neuron 2, etc.

\subsection{2D event sensor routing and mapping scheme} \label{sec:sensor-routing-scheme}

A separate pipeline for mapping and routing is available for 1D and 2D event streams originating from sensors. It is an earlier version of the event pre-processor block described in 
\cite{Richter_etal23}.

These sensor events can be routed in an alternative event word format on the 2D routing grid buses described in Section \ref{sec:sram-aer-cam-routing-scheme}. The events are then encoded with a format indicator bit (bit 23 = 1) and five variable fields as shown in Table \ref{tab:aer-words}, the event polarity \texttt{pol}, the x and y coordinates of the event (\texttt{pixel\_x} and \texttt{pixel\_y} respectively), and the target chip displacements in the x and y directions (\texttt{dx} and \texttt{dy} respectively).

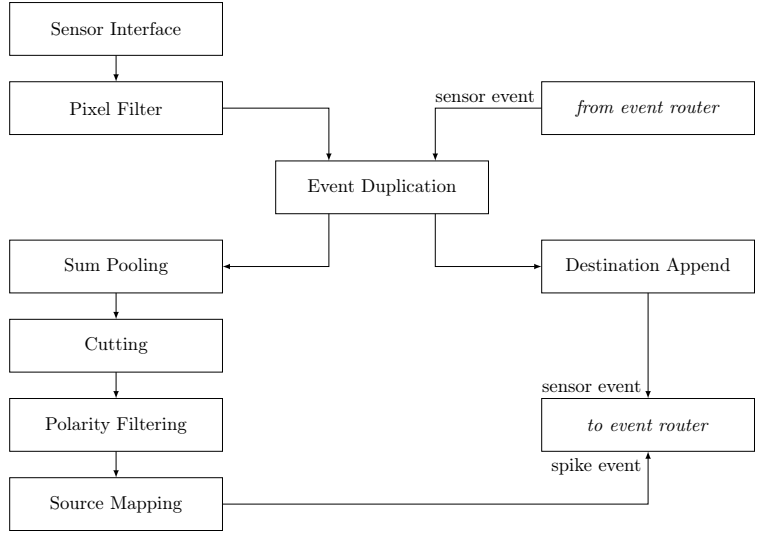
\begin{figure}[!htbp]
    \centering
    \scalebox{0.7}{
    \begin{tikzpicture}
       
        \node at(2, 9.5) {Sensor Interface};
        \draw (0,9) rectangle (4,10);
        \draw[-latex] (2, 9) -| (2, 8.5);
        \node at(2, 8) {Pixel Filter};
        \draw (0,7.5) rectangle (4,8.5);
        \draw[-latex] (4, 8) -| (6, 7);
        \draw[-latex] (10, 8) node[above left] {sensor event} -| (8, 7);
        
        \node at(7, 6.5) {Event Duplication};
        \draw (5,6) rectangle (9,7);
        
        \draw[-latex] (6, 6) |- (4, 5);
        \draw[-latex] (8, 6) |- (10, 5);
        
         \node at(12, 8) {\textit{from event router}};
        \draw (10,7.5) rectangle (14,8.5);
        
         \node at(12, 2) {\textit{to event router}};
        \draw (10,1.5) rectangle (14,2.5);
        
         \node at(12, 5) {Destination Append};
        \draw (10,4.5) rectangle (14,5.5);
        \draw[-latex] (12, 4.5) -| (12, 2.5)  node[above left] {sensor event};
        
        \node at(2, 5) {Sum Pooling};
        \draw (0,4.5) rectangle (4,5.5);
        \draw[-latex] (2, 4.5) -| (2, 4);
        \node at(2, 3.5) {Cutting};
        \draw (0,3) rectangle (4,4);
        \draw[-latex] (2, 3) -| (2, 2.5);
        \node at(2, 2) {Polarity Filtering};
        \draw (0,1.5) rectangle (4,2.5);
        \draw[-latex] (2, 1.5) -| (2, 1);
        \node at(2, 0.5) {Source Mapping};
        \draw (0,0) rectangle (4,1);
        \draw[-latex] (4, 0.5) -| (12, 1.5) node[below left] {spike event};
        
    \end{tikzpicture}}
    \caption{The sensory mapping and routing pipeline. As one pipeline can only process one patch of at most 64$\times$64 pixels, the Event Duplication block can clone and send sensor events to a second pipeline on one of the four surrounding chips by providing the target coordinates via the Destination Append block. The spike events can be sent to $\pm$7 in $x$ and $y$ chip grid coordinates.}
    \label{fig:dvs-pipeline}
\end{figure}

The mapping pipeline consists of multiple stages as shown in Fig.~\ref{fig:dvs-pipeline}. The pipeline has the following blocks:
\begin{itemize}
    \item Sensor Interface: The chip can interpret event formats from the following sensors directly via parallel AER: DAVIS346 \cite{Taverni_etal18}; DAVIS240 \cite{Brandli_etal14}; DVS128\_PAER \cite{Lichtsteiner_etal08}. Other sensors such as AEREAR2 \cite{Liu_etal14a} or ATIS \cite{Posch_etal10} can be interfaced to the event routing grid by following the sensor event word format described above.
    \item Pixel Filtering: Up to 64 arbitrary addresses can be discarded from the sensor event stream. This is done in one step using content addressable memory.
    \item Event Duplication: The pipeline can optionally duplicate and forward unprocessed events to one of the four surrounding chips.
    \item Sum Pooling: This can be used to scale the 2D input space by 1:1, 1:2, 1:4 or 1:8 in the x and y directions individually.
    \item Cutting: Cutting can be used to cut a 1$\times$1 up to 64$\times$64 pixel sized patch out of the 2D input space that is forwarded for source mapping.
    \item Polarity Filtering: Polarity selection provides the ability to use a specific polarity or both polarities.
    \item Source Mapping: A patch of 64$\times$64 pixels can be arbitrarily mapped one to one (specifying \texttt{tag}, \texttt{dx}, \texttt{dy} and \texttt{cores}) to the standard event word format. Such mapped events are introduced to the top level router for further routing and mapping inside the normal event system, as described in Section \ref{sec:sram-aer-cam-routing-scheme}.
\end{itemize}

\section{Chip interfaces} \label{sec:interfaces}



\begin{figure}
    \centering
    \includegraphics[angle=-90,trim= 820 440 0 290, clip,width=.5\linewidth]{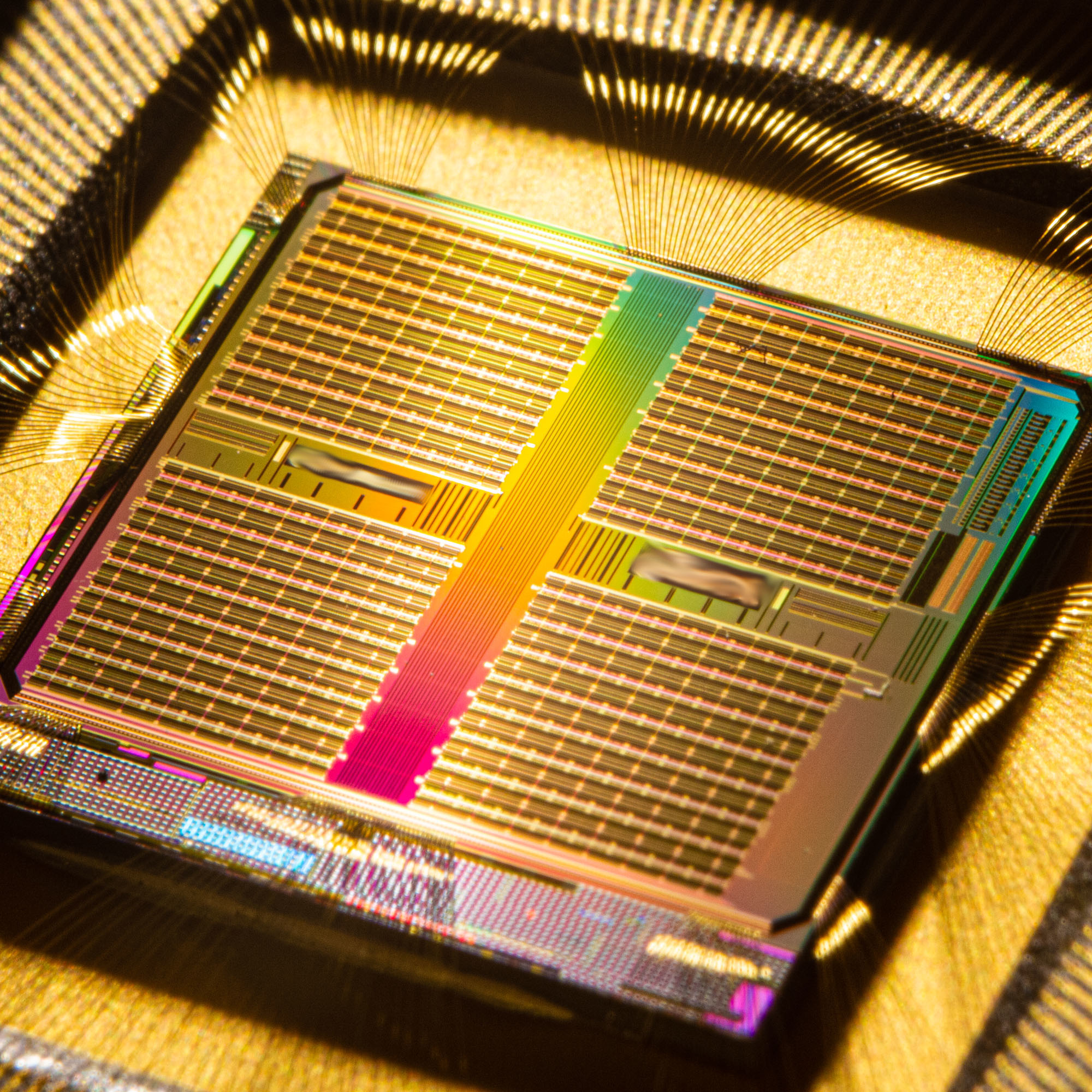}
    \caption{The corner of the chip, showing the eight channels of the Analog Front-End (AFE), including everything presented in 
    \cite{Sharifshazileh_etal21}
    .}
    \label{fig:eight-afe-channels}
\end{figure}

In addition to the North, South, East and West grid bus interfaces already alluded to in Sec.~\ref{sec:digital-spike-routing-scheme} and further remarked upon in Sec.~\ref{sec:inter-chip-event-communication} below, each \chipname{} has a multi-purpose input interface (Sec.~\ref{sec:input-interface}), a few pins for limited direct monitoring of internal signals (Sec.~\ref{sec:direct-monitoring}), outputs from on-chip monitoring circuits (Sec.~\ref{sec:on-chip-monitoring}), a sensor interface as described in Sec~\ref{sec:sensor-routing-scheme}, and inputs and monitoring points for the Analog Front-End (AFE).
See 
\cite{Sharifshazileh_etal21} 
for more details of the AFE and Fig.~\ref{fig:eight-afe-channels} for how it appears on the chip.

\subsection{Multi-purpose input-interface} \label{sec:input-interface}

The input interface (II) uses split-parallel AER to cover a wide range of configuration and communication functions including direct event input from a host system and chip configuration which in turn includes  parameter generator configuration, neuron latch configuration, connectivity memory (SRAM, CAM) configuration and natural signal AFE configuration.
Split-parallel AER means that notional 40-bit AE words are presented in two cycles of $40/2 + 1$ bits each, i.e.~one cycle for the most significant and one for the least significant half of each AE word.
The additional bit in each cycle is used to differentiate the two half-words.
This split parallel operation is chosen to keep the chip pin count more manageable.

\subsection{Inter-chip event communication} \label{sec:inter-chip-event-communication}

As alluded to in Sec.~\ref{sec:digital-spike-routing-scheme}, 
each chip has four high-speed asynchronous AER buses on the four sides to directly transfer events in and out of the chip. The pins are assigned in such a way that adjacent chips can be conveniently connected together, which facilitates network scalability across a 2D grid. Each chip can directly address a neighborhood comprising up to seven chips in each direction, which allows a maximal $8\times8$ fully connected chip array without any external mapping.
Using an alternative packet format, this grid also transmits and receives sensor events to and from its direct neighbors, see Sec.~\ref{sec:sensor-routing-scheme}.

\subsection{Direct monitoring} \label{sec:direct-monitoring}

Some important analog signals are copied to six external pins through rail-to-rail buffers so that they can be monitored directly off-chip using an oscilloscope for debugging purposes. These are a neuron membrane potential from all four cores and analog voltage or current reference parameters from Parameter Generators 0 and 1.
Also externally available are digital homeostasis charging direction signals from all four cores and a digital delay pulse extender pulse from particular synapses on each core.

\subsection{On-chip monitoring} \label{sec:on-chip-monitoring}

Sixty-four on-chip, current-based spiking analog to digital converters (sADC) ensure easy monitoring of all relevant neural signals. This greatly improves the configuration experience and usability.

The signals are divided into three separately configurable groups, in order to adapt to the wide range of signal magnitudes.

\begin{table}[!htbp]
    \centering
    \caption{sADC groups}
    \begin{tabular}{ccc}
        \hline
        Group 0 & Group 1 & Group 2 \\ \hline
        External profiling voltage & Internal calibration voltage & Internal calibration voltage \\
        Membrane potential & Adaptation/Ca DPI & Synapse 20/41 weight \\
        Refractory pulse extender & Dendritic DPIs & Homeostasis gain \\ \hline
    \end{tabular}
    \label{tab:sadc-groups}
\end{table}


\section{Software} \label{sec:software}

\subsection{\companyname{} \softwarename{}}

In order to enable users to perform experiments with \chipname{} chips, software support is provided through the \softwarename{}\footnote{name hidden for double-blind review process} software \cite{\softwarename{}}.
\softwarename{} has been developed by \companyname{} \cite{\companyname{}} 
to support a diverse range of current and future neuromorphic chips.



\subsubsection{Objectives of \softwarename{}}

\softwarename{} aims to: provide unified support for a diverse range of neuromorphic chips; be `remotable'; provide a GUI, and at the same time a conventional programming interface; be performant; and run on multiple operating systems.

All of the chips supported by \softwarename{} should be supported in a similar way, such that once a user is familiar with the GUI and the API for one chip, the experience is reasonably portable to the use with other chips, thus saving the user familiarization time.

\softwarename{} aims to support the remote use of \chipname{} (and other chips) such that the user interface and user-supplied code can (but need not) run on a different computer (e.g.~the user's laptop) from the computer to which the chips are attached, be it at the same desk, in a server room in the same building, or half-way around the world.
This facility has already proved invaluable in teaching.
Students working at home have been able to perform experiments on \chipname{} chips without having to be physically provided with the hardware.

Experience with earlier generations of mixed-signal neuromorphic chips has shown that it is highly advantageous to provide a graphical user interface (GUI) to provide visual feedback of, for instance, neural spiking activity, and to provide on-screen virtual potentiometers to control on-chip analog parameters.
This is particularly important while initially tuning those parameters.
At the same time, for anything beyond this most trivial of interactions, an application programming interface (API) of some kind is essential.
In earlier software, the existence of these two interfaces, through which the state of the neuromorphic chips which are being used could be altered, caused problems, as the state could be changed in the GUI without this being apparent to code using the API and {\it vice versa}.
Avoiding these kinds of discrepancies between different components' view of the state information has been key to the architecting of \softwarename{}.

The API presented to the user is in Python 3, as Python has become the {\it de facto} standard in neuroscience, in particular in the field of modeling and simulation \cite{Muller_etal15, Davison_etal09}.
The underlying code, however, is in C++ (C++17) for performance.

Finally, for broad acceptance and ease of use, it is important that \softwarename{} is supported on multiple platforms. Currently Linux and macOS are supported.
 
\subsection{The Software and Hardware Stack}

\begin{figure}[!htbp]
    \centering
    \begin{tikzpicture}
        \node[double arrow, shape border rotate=90, minimum height=1.5cm, draw] (usb) {};
        \node[anchor=west] at (usb.east) {USB};
        \node[draw, anchor=south, minimum width=10cm] (libusb) at (usb.north) {libusb};
        \node[rectangle, draw, anchor=south, minimum width=12cm, minimum height=10cm] (\softwarename{}) at (libusb.north) {};
        \node[anchor=north west] at (\softwarename{}.north west) {\softwarename{}};
        \node[rectangle split, rectangle split parts=2, draw, minimum width=10cm, text centered, anchor=south] (uf-ok) at (libusb.north) {FPGA firmware communication module \nodepart{two} OpalKelly module};
        \node[anchor=north east] at (uf-ok.north east) {*};
        \node[rectangle split, rectangle split parts=2, draw, minimum width=10cm, text centered, anchor=north] (OK-fpga) at (usb.south) {OpalKelly XEM73x0 \nodepart{two} FPGA firmware};
        \node[rectangle, draw, above=3.5cm of OK-fpga, minimum width=10cm, minimum height=4cm] (d2) {};
        \node[anchor=north west] at (d2.north west) {\texttt{\chipname{}} module};
        \node[anchor=north east] at (d2.north east) {*};
        \node[draw, above=0.5cm of d2] (ros) {Remote object store};
        \node[draw, above=0.5cm of ros, xshift=-2cm] (rl) {Remoting layer};
        \node[align=center, draw, above=0.5cm of \softwarename{}, xshift=-2cm, minimum height=1.5cm, minimum width=2cm] (upc) {User's \\ Python \\ code};
        \node[draw, above=1.5cm of ros, xshift=2cm, minimum height=1.5cm, minimum width=2cm] (gui) {GUI};
        \node[draw, above=6cm of OK-fpga, xshift=-2.5cm] (d2i) {\texttt{\chipname{}Interface}};
        \node[draw, above=6cm of OK-fpga, xshift=2cm, minimum width=4cm] (d2m) {\texttt{\chipname{}Model}};
        \node[align=center, anchor=north east, draw] at (d2m.south east) (d2c) {\texttt{\chipname{}} \\ \texttt{Configuration}};
        \node[draw, above=4cm of OK-fpga, xshift=2cm] (evgen) {Configuration event generators};
        \draw[<->] (upc) -- (rl);
        \draw[<->] (gui) -- (ros);
        \draw[<->] (rl) -- (ros);
        \draw[<->] (ros) -- (d2i);
        \draw[<->] (ros) -- (d2m);
        \draw[{Circle[]}-{Circle[]}] (d2i.east) -- (d2m.west);
        \draw[<->] (d2i.south) -- (uf-ok);
        \node[right=0.5cm of d2m.south west] (d2m-ssw) {};
        \node[right=1cm of evgen.north west] (evgen-nnw) {};
        \draw[->] (d2m-ssw.base) -- (evgen-nnw.base);
        \node[right=1cm of evgen.south west] (evgen-ssw) {};
        \draw[->] (evgen-ssw.base) -- (uf-ok);
        \node[align=center, rectangle split, rectangle split parts=4, rectangle split horizontal, draw, text width=2.25cm, text centered, anchor=north] (chips) at (OK-fpga.south) {\chipname{} \\ chip 0 \nodepart{two} \chipname{} \\ chip 1 \nodepart{three} \chipname{} \\ chip 2 \nodepart{four} \chipname{} \\ chip 3 };
    \end{tikzpicture}
    \caption{
    Full software and hardware stack showing the main \chipname{} related components of \softwarename{} and the corresponding hardware.
    The modules marked with a * were written as part of the \chipname{} project.
    When \softwarename{} is used with other chips, the corresponding module (not shown) is used instead of the \texttt{\chipname{}} module.
    If a different interface board is used, a different module will be used instead of the OpalKelly module.
    }
    \label{fig:sw-hw-stack}
\end{figure}
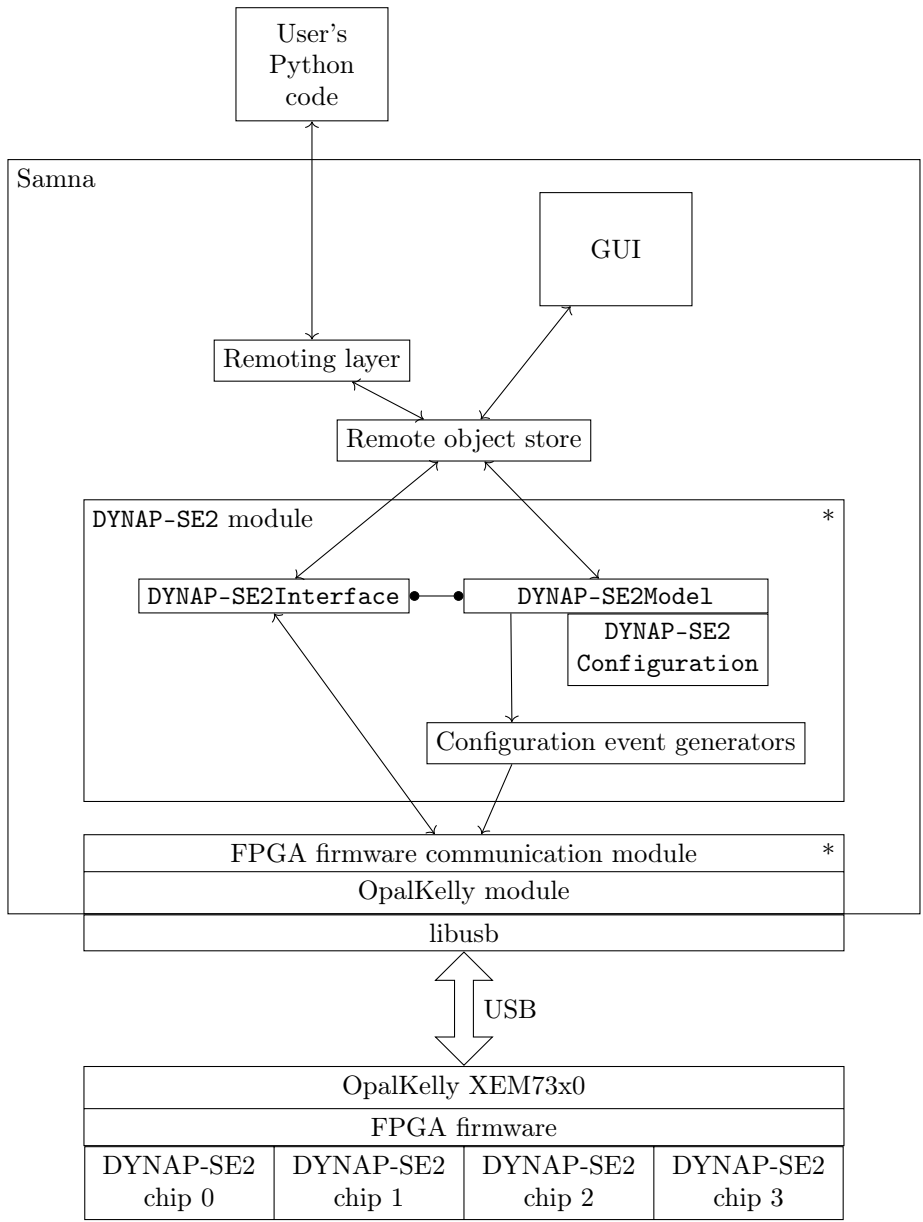

Figure \ref{fig:sw-hw-stack} shows the full stack of \chipname{} software and hardware, from the user's Python code and the GUI at the top to the \chipname{} chips at the bottom.

The following description concentrates on the \chipname{} and FPGA firmware communication modules of \softwarename{}, as these were the modules written in the course of the \chipname{} project.

\subsubsection{User code, GUI and object store}

Although the GUI is part of \softwarename{}, it is on equal footing with the user's code when accessing the rest of the system.
Both talk to the \chipname{} module of \softwarename{} via a local remoting layer and a remote object store where the remote object store and everything below it may be on a remote computer.
Objects from the \chipname{} module (and other similar modules supporting other hardware, not shown in Fig.~\ref{fig:sw-hw-stack}) can be placed in the object store and transparently retrieved from there by the user's code and/or the GUI.
They can then be manipulated and returned to the store and thus to the lower modules.

The user's Python code only sees a Python extension library which can be imported in the usual fashion:
\begin{verbatim}
import <software>
from <software>.<chip> import *
\end{verbatim}
\begin{verbatim}
import samna
from samna.dynapse2 import *
\end{verbatim}
From this point on, barring a little setup to connect to a remote \softwarename{} node, the user need not be aware of the presence of the object store, or that the hardware might be attached to a remote machine.
The classes in \verb+<software>.<chip>+
\verb+samna.dynapse2+ can all be used transparently as if everything was local.
Within \softwarename{}, the actual Python interface to the underlying C++ code is implemented with the aid of pybind11 \cite{Jakob_etal17}.

\subsubsection{\chipname{} module}
\label{\chipname{}-module}

Within \softwarename{}'s \chipname{} module, there are \chipname{}Interface classes which provide an interface to facilities provided by the PCB(s) on which the \chipname{} chips are mounted.
At the time of writing, two \chipname{}Interface classes exist: \chipname{}DevBoard and \chipname{}Stack are available for the two present PCB types, \textsc{dev board} and \textsc{stack}
respectively.
Alongside the \chipname{}Interface class is the \chipname{}Model class which provides an interface to an abstraction, held in a \chipname{}Configuration class, of the hardware state in the physical \chipname{} chip(s).
The \chipname{} chips do not support the read-out of internal state, so the entire state information is held in software in the \chipname{}Configuration class and other aggregated classes which are not shown in the figure.
See Sec.~\ref{\chipname{}Configuration-hierarchy} below for details.

\begin{figure}[!htbp]
    \centering
    \begin{tikzpicture}
        \draw (0, 4) node[rectangle, draw] (get-model) {\texttt{model = interface.get\_model()}};
        \draw (0, 3) node[rectangle, draw] (get-config) {\texttt{cfg = model.get\_configuration()}};
        \draw (0, 2) node[rectangle, draw] (modify) {Modify \texttt{cfg} as desired};
        \draw (0, 1) node[rectangle, draw] (apply-config) {\texttt{model.apply\_configuration(cfg)}};
        \draw[->] (0, 5) -- (get-model.north);
        \draw[->] (get-model.south) -- (get-config.north);
        \draw[->] (get-config.south) -- (modify.north);
        \draw[->] (modify.south) -- (apply-config.north);
        \draw[->] (apply-config.south) -- (0, 0) -- (-4, 0) -- (-4, 3) -- (get-config.west);
    \end{tikzpicture}
    \caption{\texttt{get\_configuration()}, modify configuration, \texttt{apply\_configuration()} loop.}
    \label{fig:get-config-apply-config-loop}
\end{figure}
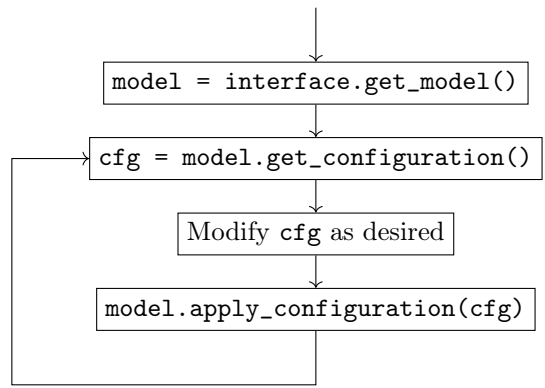

In operation, the user's code, and/or the GUI, obtains a reference to a \chipname{}Model object via the object store, then gets the current configuration of the hardware from the \chipname{}Model object as a \chipname{}Configuration object, modifies that object and the tree of objects within it representing the neuron and synapse configuration information, and applies the \chipname{}Configuration object back into the \chipname{}Model object and hence to the hardware.
This process can then be performed repeatedly, see Fig.~\ref{fig:get-config-apply-config-loop}.
In this way, changes made by the user's code are visible to the GUI and {\it vice versa}.

When the \chipname{}Configuration object is set back into the \chipname{}Model object, the \chipname{}Model determines the changes from the current configuration and uses event generator functions to produce a list of configuration events sufficient to bring about those changes on the \chipname{} chip(s).
This list of events is then passed to the \chipname{}Interface object for transmission to the hardware.
Meanwhile, address-event (AE) streams to and from the hardware pass directly to and from the user code and the GUI directly via the same \chipname{}Interface object.

\subsubsection{FPGA firmware communication module and below}

The FPGA firmware communication module manages the packet-based communication with the firmware instantiated in the FPGA on the hardware.
To avoid overhead associated with constantly allocating and freeing packet buffers, the firmware communication module manages a pool of constant-length packet buffers.
Empty packet buffers are then obtained by the overlying hardware-specific module(s), in this case by a \chipname{}Interface object in the \chipname{} module, when there are events to send to the hardware.
The hardware-specific module is responsible for filling in the payload of the packet before calling back into the firmware communication module to let the latter complete the header of the packet with appropriate payload size information and put the packet on a transmit queue.

The firmware communication module is also home to a thread which continually attempts to read from the underlying hardware platform support module.
At the time of writing, for \chipname{} this is always the OpalKelly module, since both the supported \textsc{dev board} and \textsc{stack} boards interface via Opal Kelly \cite{OpalKelly} FPGA Integration Modules.
After each read, the firmware communication module determines whether the firmware is ready to accept more data, and if so, how much.
It then takes as many packets as possible from the transmit queue and writes them out via the OpalKelly module, packing them into the blocks that the OpalKelly layer understands.
Once the packet buffer contents have been copied into the OpalKelly blocks, the packet buffers are returned to the packet buffer pool.

The OpalKelly model abstracts the `Pipe' and `Wire' interface provided by the Opal Kelly hardware and communicates with the hardware via libusb \cite{libusb}.
Finally when the Opal Kelly board receives the blocks assembled by the software, the FPGA firmware unpacks the individual packets from the Opal Kelly blocks and passes 
the event data contained in the packets to the \chipname{} chips via the appropriate bus.

\subsubsection{Events from the chip(s)}

Events coming from the inter-chip communication buses
and from the sADC output of the \chipname{} chip(s) are read by the firmware in the FPGA on the Opal Kelly board.
In the case of inter-chip communication events, these events are timestamped and placed in packets.
In the case of sADC events, the number of events received for each possible sADC address in a fixed time interval are counted, and all of these counts are placed into a different packet type at the end of the interval.
In both cases, the packets are placed in blocks and transmitted over USB to the host.
When these blocks are read by the thread referred to above in \softwarename{}'s FPGA firmware communication module, the packets are unpacked from the blocks into buffers taken from the packet buffer pool and dispatched according to packet type.
In the case of the normal timestamped events, the packets are placed into a queue from which they can be read by the top-level code via the \chipname{}Model object.
In the case of sADC count packets, the packet contents are written into a buffer which always holds the latest sADC count values which is also available to be read by top-level code via the \chipname{}Model object.

\subsubsection{\chipname{}Configuration aggregation hierarchy}
\label{\chipname{}Configuration-hierarchy}

As mentioned above in Sec.~\ref{\chipname{}-module}, the entire \chipname{} hardware state information is held in the software in \chipname{}Configuration objects and a hierarchical aggregation of Plain Old Data (POD) types and objects of further classes: \chipname{}Chip, \chipname{}Core, \chipname{}Neuron, \chipname{}Synapse etc., which themselves are (almost all) POD types, i.e.~they are aggregates with only public data.
It is this hierarchically organised data structure which the user manipulates in their Python code to control the operation of the \chipname{} chips.

\section{Discussion}

The large range of dynamics and computing features supported by the \chipname{} support the definition of networks that can solve a wide range of applications.
Similarly, the \chipname{} fully configurable tag-based routing system enables the definition of arbitrary network topologies, ranging from simple feed-forward neural networks, to fully recurrent ones.

Feed-forward networks are the simplest form of network architectures, in which the neurons process events as they move through the layers of the network.
Sparse feed forward networks can be built by dividing the available neurons into layers, and forming unidirectional synaptic connections between layers~\cite{Frenkel21}.
Unlike in standard crossbar and addressable column approaches~\cite{Qiao_etal15, Merolla_etal14a}, the CAM-based synaptic addressing allows all the available physical synapses to be used~\cite{Moradi_etal18}.
To support dense feed forward networks and allow users to define heterogeneous networks with different fan-in and fan-out figures, each core allows the number of programmable synapses to be increased to 256 per neuron, at the cost of a reduced number of  neurons (64 instead of 256). 

The asynchronous and mixed signal design of the \chipname{} is particularly well suited for emulating the dynamics of recurrent spiking neuronal network architectures.
The native support for recurrent mapping and continuous physical time emulation overcomes the limits of digital time-multiplexed simulation systems, avoiding the need for complex clock tree designs and reducing signal synchronization issues.
Reservoir networks use recurrent connections to build complex network dynamics supporting a `memory trace' of their activity over time.
Attractor networks can exploit recurrent connectivity patterns to memorize patterns, recover partial or corrupted input patterns, and perform stateful computation~\cite{Cotteret_etal22, Liang_Indiveri19}.

Both feed-forward and recurrent networks can be configured to implement time-to-first-spike (TTFS) computation. This paradigm relies on the latency of spike waves traveling through a network, like wavefront algorithms~\cite{Ponulak_Hopfield13} or as seen in the nervous systems of weakly electric fish~\cite{Engelmann_etal16}.
The low-latency nature of \chipname{} and its ability to support delay-based synapses make TTFS applications first class citizens.
In particular, the fact that synapses can be configured to belong to one of four delay classes (with two well-matched -precise-  classes, and two purposely mismatched -inhomogeneous- classes)  provide a controlled distribution of delays which enables both precise time-to-first-spike configurations, and randomly timed networks~\cite{George_Indiveri16,Sheik_etal12}.

The ability to configure synapses as diffusive gap junctions~\cite{Neckar_etal19} with 2D nearest neighbor connections supports the configuration of networks with local spatially distributed connectivity kernels, as originally proposed in~\cite{Benjamin_etal21,Benjamin_etal14}.
In addition, excitatory synapse circuits can be configured to emulate both slow voltage-gated NMDA receptor dynamics~\cite{Bartolozzi_Indiveri07a} as well as fast AMPA dynamics~\cite{Sumislawska_etal16}.
For both AMPA and NMDA synapse types (as well as both inhibitory types, GABA-A and GABA-B), the 4-bit weight resolution, combined with the configurable weight-range scale enable users to explore and implement more advanced hardware-in-the-loop learning systems.

The improved spike-frequency adaptation circuits present in the neuron circuits~\cite{Rubino_etal20}, the neuron's homeostatic synaptic scaling circuit~\cite{Qiao_etal17}, and the synapse short term depression plasticity control~\cite{Bartolozzi_Indiveri07a} provide the user with a large range of computational primitives for exploring dynamics at multiple time scales and produce complex dynamic behaviors~\cite{Jaeger_etal21}.

Finally, the ability to monitor all dendritic, somatic and synaptic current traces via asynchronous current-to-frequency ADCs~\cite{Yang_etal12} greatly simplifies prototyping and debugging in experiments that explore the dynamics and computing abilities of the \chipname{}.

\section{Conclusion}

We presented a full custom implementation of a 
\chipname{},  built for prototyping small networks of spiking neurons that emulate the dynamics of real neurons and synapses with biologically plausible time constants, for interacting  with natural signals in real time.
We argued that the real-time nature of the system and its direct sensory-input interfaces for receiving 1D and 2D event streams make this an ideal platform for processing natural signals in closed-loop applications.
We characterized in detail all circuits present on the chip and presented chip measurements that demonstrate their proper operating function.
This platform will enable the prototyping of biologically plausible sensory-processing systems and the construction of physical neural processing systems that can be used to validate (or invalidate) hypotheses about neural computing models.


\ack{
The authors would like to thank Mohammadali Sharifshazileh his work on the initial PCB design and management of the packaging process.
The authors would like to thank Melika Payvand for her help with design submission administration and additional simulations of the AMPA resistive grid design.
The authors would like to acknowledge Sunil Sheelavant for his contribution to the pad frame schematic.\\

The authors would like to acknowledge the financial support from SynSense AG, Thurgauerstrasse 60, 8050 Zurich, Switzerland, and NeuroAgents an ERC Consolidator project (No. 724295) https://doi.org/10.3030/724295 .
Ole Richter would like to acknowledge the financial support of the CogniGron research center and the Ubbo Emmius Funds (University of Groningen). During the ASIC design Ole Richter was solely affiliated to SynSense AG.

\textbf{Author Contribution:}

G.I. conceived the original concepts and analog circuit designs. Q.N. and O.R. implemented the mixed-signal and the asynchronous circuit designs and architecture. C.W. and G.K. designed the printed circuit board and peripheral logic. C.W. and C.N. wrote the low-level firmware. A.W. and C.N. wrote the high-level software. C.W. carried out chip measurements and testing. All authors contributed to the manuscript writing efforts, G.I. supervised the project.
}
\printbibliography

\end{document}